\documentclass[a4paper,11pt]{article}

\usepackage{amsmath,amsfonts}
\usepackage{algorithmic}
\usepackage{algorithm}
\usepackage{array}
\usepackage{authblk}
\usepackage[caption=false,font=normalsize,labelfont=sf,textfont=sf]{subfig}
\usepackage{textcomp}
\usepackage{stfloats}
\usepackage{url}
\usepackage{verbatim}
\usepackage{graphicx}
\usepackage{threeparttable}
\usepackage{blkarray}
\usepackage{tikz}
\usepackage{float}
\usepackage{enumitem}

\newcommand{\circo}{\raisebox{-0.5pt}{\tikz \draw[line width=0.8pt] circle(3pt);}}
\newcommand{\diamondo}{\raisebox{-0.32pt}{\tikz \draw[line width=0.8pt] (0,0) -- (0.12,0.12) -- (0,0.24) -- (-0.12,0.12) -- cycle;}}
\newcommand{\triangleo}{\raisebox{-0.2pt}{\tikz \draw[line width=0.8pt] (0,0) -- (0.12,-0.2) -- (-0.12,-0.2) -- cycle;}}
\newcommand{\squareo}{\raisebox{-0.25pt}{\tikz \draw[line width=0.8pt] (-0.1,-0.1) rectangle (0.1,0.1);}}

\usepackage{lineno,hyperref}
\usepackage{multirow}
\usepackage{booktabs}
\usepackage{ulem}
\usepackage{mathptmx}
\usepackage{amsthm}
\usepackage{dsfont}
\usepackage{color}
\usepackage{tikz}
\usepackage{pgfplots}
\usetikzlibrary{calc}
\usepackage{caption}
\usepackage{todonotes}

\newtheorem{proposition}{Proposition}

\newtheorem{definition}{Definition}

\newcommand{\R}{\mathds{R}}
\newcommand{\argmin}{\mathop{\mathrm{argmin}}}

\newcommand{\x}{{\mathbf{x}}}            
\newcommand{\y}{{\mathbf{y}}}            
\newcommand{\e}{{\mathbf{e}}}            
\newcommand{\bu}{{\mathbf{u}}}           
\newcommand{\ba}{{\mathbf{a}}}           
\newcommand{\bb}{{\mathbf{b}}}           

\begin{document}

\title{Absolute indices for determining compactness, separability and number of clusters}


\author[1]{Adil M. Bagirov}
\author[2]{Ramiz M. Aliguliyev}
\author[1]{Nargiz Sultanova}
\author[3]{Sona Taheri}

\affil[1]{Centre for Smart Analytics, Institute of Innovation, Science and Sustainability, Federation University Australia, Ballarat, Australia}
\affil[2]{Institute of Information Technology, Baku, Azerbaijan}
\affil[3]{Mathematical and Geospatial Science, RMIT University, Melbourne, Australia}

\maketitle

\begin{abstract}
Identifying ``true" clusters in a dataset is inherently challenging. Clustering models and algorithms may fail to produce compact, well-separated groups or to correctly determine the optimal number of clusters. Cluster validity indices are often applied to find such clusters. However, most existing indices are relative measures, designed primarily for comparing clustering algorithms or tuning their parameters. In this paper, we introduce new absolute cluster validity indices that assess both cluster compactness and separability. For each cluster, we define a compactness function, and for each pair of clusters, we define an associated set of neighboring points. The compactness function measures the cohesion of individual clusters as well as the entire clustering distribution, while the neighboring-point sets are used to define the margin between cluster pairs and the overall distribution margin. These compactness and separability indices are then employed to estimate the true number of clusters. We evaluate the proposed indices on a variety of synthetic and real-world datasets and compare their performance with several widely used cluster validity measures.
\end{abstract}


\section{Introduction}
Clustering is a fundamental task in data mining with a wide range of practical applications, including medicine and bioinformatics \cite{Mustjoki2017}, economics  \cite{Mardaneh2016}, cybersecurity  \cite{Taheri2020cyber}, trajectory representation learning \cite{meng2025survey}. Determining the optimal or ``true” number of clusters in a dataset remains a highly challenging problem. As noted in \cite{Jain1988}, ``refining clustering processes to obtain accurate, meaningful results is an art". There is no universally accepted criterion for identifying the optimal number of clusters. In practice, this number is typically inferred from cluster distributions having compact and well-separated clusters, yet such distributions may not be unique for a given dataset.

Cluster validity indices are commonly employed to estimate the appropriate number of clusters. In \cite{bagkartah2024,kaufman1990}, various validity indices for cluster validation are presented. The Silhouette index, which remains one of the most widely used internal validation measures, is introduced in \cite{Rousseeuw1987}. The sum-of-squares-based criteria for assessing cluster quality is proposed in \cite{Franti2009,Zhao2014}, while a distance-based separability measure is proposed in \cite{guan2022distance} for internal cluster validation. These indices typically assess cluster compactness and/or separability. Classical validity measures -- such as the Davies-Bouldin \cite{Bouldin1979}, Calinski–Harabasz \cite{Calinski1974}, Dunn \cite{Dunn1974}, and Silhouette \cite{Rousseeuw1987} indices -- combine both intra-cluster compactness and inter-cluster separation.

Most cluster validity indices are relative in nature and serve primarily to compare clustering algorithms or fine-tune algorithm parameters. In datasets with complex clustering structures, different indices often yield conflicting recommendations \cite{Zimek2020}. As shown in \cite{Zimek2020}, absolute validity indices are generally more effective for evaluating the compactness and separability of clusters produced by a single clustering algorithm. Separability has also been explored in supervised settings where class labels are available. Examples include the nearest-neighbor-based Geometric separability index \cite{thornton1998separability}, graph-based neighborhood metrics \cite{zighed2002separability}, statistical distances like the Jeffries–Matusita distance \cite{dabboor2014jeffries}, and other separability criteria \cite{mishra2009separability,Peterson2011}. 

In this paper, we propose a novel approach to constructing an absolute cluster validity index. Existing validity measures often assess separability using distances between cluster centers or aggregate distance statistics, which may not adequately capture the actual boundaries between neighboring clusters. To address this limitation, we introduce a new separability measure based on adjacent sets and cluster margins. We also define the notion of a compactness function for a cluster, which serves as the basis for assessing cluster compactness. The compactness of the cluster distribution is derived from the compactness of all clusters, while the margins between neighboring clusters are used to quantify pairwise separability and, subsequently, the separability of the overall cluster distribution. Both compactness and separability criteria are employed to determine the number of clusters. The proposed indices are evaluated on several synthetic and real-world datasets, and their performance is compared with a number of well-known cluster validity indices.

The notion of cluster compactness, introduced in this study, is inspired by the mathematical concept of compactness. The separability index of cluster distributions is determined using those data points that are most relevant to distinguishing between clusters. In many real-world datasets, cluster compactness and separability are competing objectives, creating a trade-off that motivates the formulation of a multi-objective optimization model for defining a cluster validity index. Accordingly, the cluster validity index developed in this paper is fundamentally different from existing validity indices, including the absolute cluster validity index proposed in \cite{Zimek2020}.

The remainder of the paper is organized as follows.  Section \ref{related} provides an overview of the related work. The compactness function and index are defined in Section \ref{defnot}. An index for expressing the separability of clusters is presented in Section \ref{separable}. The problem of determining the number of clusters is discussed in Section \ref{optimalnum}, followed by numerical experiments in Section \ref{numerical}. Finally, Section \ref{conclusions} provides discussion of results and concluding remarks.

\vspace{-3mm}

\section{Related work} \label{related}
Since the early days of cluster analysis, identifying the optimal number of clusters has remained a significant challenge \cite{Hennig2015, Milligan1985}. This number corresponds to cluster configurations in which clusters are highly compact and well separated, which in turn requires precise definitions of ``compactness” and ``separability.” However, real-world datasets often exhibit irregular point distributions arising from noise, outliers, or other factors. Consequently, cluster compactness and separability can vary significantly between data, and a single criterion is unlikely to provide a universally valid definition of the optimal number of clusters.

Objective functions in most optimization-based clustering models are designed to capture intra-cluster similarity and inter-cluster dissimilarity. The optimal values of these functions can be considered as internal criteria for assessing clustering quality, particularly with respect to compactness. However, they do not always provide an effective measure of cluster separability as they typically assess separation using cluster centers or aggregate distances. Examples of such approaches include parametric modeling a quantization error \cite{Kolesnikov2015}, the use of internal cluster validity indices \cite{Zhou2018}, and a curvature-based algorithm \cite{Zhang2017}. In addition, \cite{Saha2021} proposed an algorithm that identifies the number of clusters during the clustering process.

There have also been efforts to incorporate cluster validity indices directly into clustering models. For example, \cite{Batool2021} used the average silhouette width as an objective function to obtain well-separated clusters, and \cite{BagAlSu2023} employed the silhouette index to develop a model and algorithm to find compact and well-separated clusters. In \cite{BagirovCAOR} an optimization model was introduced specifically to improve cluster separability.

Cluster validity indices are primarily used to assess the quality of clustering solutions by examining two key aspects: compactness and separability \cite{Bouldin1979,Dunn1974,Rousseeuw1987}. Several comparative studies  investigated the behavior and limitations of cluster validity indices. For example, in  \cite{Franti2009}, the statistical significance of sum-of-squares-based validity measures was analyzed including comparative assessment against other cluster validity indices. An experimental comparison of internal validity indices was conducted in \cite{Perez2012}. A systematic overview and critical assessment of clustering benchmarks and evaluation methodologies were provided in \cite{Gagolewski2021}.

Additional relative indices have been introduced, extending classical compactness and separability criteria. For example, \cite{halkidi2001clustering} integrates compactness and separability by measuring cluster variance and inter-cluster density; \cite{cheng2018novel} introduces a local core-based method that leverages graph-based distances between dense regions to capture cluster structure; and \cite{liang2020cluster} proposes an index tailored for irregular or non-convex clusters, which accounts for density variations within and between clusters to evaluate clustering quality in datasets with heterogeneous densities. More recently, the Multi-Granularity Fusion index was proposed in  \cite{tang2025clustering} to improve the robustness of fuzzy cluster validity assessment in noisy and high-dimensional datasets. A correlation-based index is introduced in \cite{WIROONSRI2024109910} to identify multiple plausible cluster numbers.

Most cluster validity indices are applied as post-clustering evaluation tools rather than being incorporated directly into the clustering process. An absolute cluster validity index based on a geometric characterization of the solution space was proposed in \cite{Zimek2020}. This index offers a coherent description of the data structure by jointly integrating inter- and intra-cluster distances, cluster density, and multimodality within clusters.

No cluster validity measure can be entirely independent of the underlying data structure, however, the indices proposed in this paper aim to reduce this dependence by employing absolute measures of compactness and separability. The compactness index is based on a normalized compactness function, while the separability index evaluates neighboring clusters through adjacent sets and cluster margins rather than relying solely on distances between cluster centers. As a result, the proposed framework focuses on intrinsic properties of the cluster distribution and provides a more interpretable assessment of clustering quality across datasets with different characteristics. Combined, the proposed compactness and separability indices form an absolute cluster validity framework for determining the number of clusters.

The main contributions of this study are:
\begin{itemize}[nosep]
\item A novel compactness index is proposed for evaluating clustering quality. The index is based on a compactness function that characterizes the distribution of distances between data points and their corresponding cluster centers;
\item A new separability index is introduced based on the concepts of adjacent sets and cluster margins. The proposed index explicitly quantifies the separation between neighboring clusters through their boundary structure;
\item The compactness and separability measures are combined into an absolute cluster validity framework. Both indices are scaled and independent of the ordering of observations and attributes;
\item The determination of the number of clusters is formulated as a multiobjective optimization problem, and decision-space plots for selecting optimal clustering solutions are introduced;
\item Computational experiments on synthetic and real-world datasets are conducted, and the performance of the proposed indices is compared with several established cluster validity measures.
\end{itemize}

\vspace{-3mm}

\section{Compactness function and index} \label{defnot}
In this section, we define the compactness function and introduce the compactness index. First, we recall the definition of the hard clustering problem. Let $k \geq 2$ be a given integer, and let $A=\{\ba_1,\ldots,\ba_m\} \subset \R^n$ be a finite set of points, where $\R^n$ denotes the $n$-dimensional Euclidean space, the points $\ba_i,~i = 1,\ldots,m$ are referred to as observations, and each point has $n$ attributes. The goal of hard clustering is to partition $A$ into $k$ nonempty, disjoint subsets (clusters) $A_j,~j=1,\ldots,k$, using a predefined similarity measure, such that $\cup_{j=1}^k A_j = A.$ 

\subsection{Compactness function} 
Consider the set $A$. Let $d: \R^n \times \R^n \rightarrow \R_+$ be a similarity measure defined using a distance function, and let $\x$ be the center of $A$ with respect to the similarity measure $d$, that is
$$
\x = \argmin_{\y \in \R^n} \Big\{\frac{1}{m}\sum_{\ba \in A} d(\y,\ba) \Big\}.
$$
The radius $R_A$ and the average radius $R_{av}$ of $A$ are defined as:
\begin{equation} \label{radius}
R_A = \max_{\ba \in A} ~d(\x,\ba)~~\mbox{and}~~R_{av} = \frac{1}{|A|} \sum_{\ba \in A} d(\x,\ba),
\end{equation}
where $|\cdot|$ is the cardinality of a set. Define the set
\[
S(t) = \Big\{\ba \in A: ~d(\x,\ba) \leq t \Big\},~t \in [0, \infty).
\]
If $\x \in A$, then $S(0) = \{\x\}$; otherwise $S(0) = \emptyset$. Additionally, we have $S(R_A) = A$ and $S(t) = S(R_A)$ for all $t \in [R_A, \infty).$ Here, $t$ serves as a distance threshold defining the radius of a sphere centered at $\x$; as $t$ increases, $S(t)$ captures an expanding neighborhood of points around the center, and the compactness function $f(t)$ introduced below records the average distance of the enclosed points from $\x$.

\begin{definition} \label{scattering}
The compactness function $f: [0,\infty) \rightarrow [0,\infty)$ of the set $A$ is defined as:
\vspace{-3mm}
\begin{align}
	f(t) = \left\{
	\begin{array}{ll} \label{scatfunc}
		0, & \quad \mbox{if}~~S(t) =\emptyset, \\
		\frac{1}{|S(t)|} \sum\limits_{\ba \in S(t)} d(\x,\ba), & \quad \mbox{if}~~S(t) \neq \emptyset.
	\end{array} \right.
\end{align}
\end{definition}

\begin{proposition} \label{increasing}
The compactness function $f$ is a non-decreasing step function.
\end{proposition}
\begin{proof} Take any $t_1, t_2 \in [0, R_A]$ with $t_1 < t_2$. We show that $f(t_1) \leq f(t_2)$. Note that $S(t_1) \subseteq S(t_2)$. If $S(t_1)=\emptyset$, then $f(t_1) \leq f(t_2)$. If $|S(t_1)|=|S(t_2)|$, we get $f(t_1)=f(t_2)$. Assume that $S(t_1) \neq \emptyset$ and $S(t_1) \neq S(t_2)$. Denote by $\hat{S} = S(t_2) \setminus S(t_1)$. Then $\hat{S} \neq \emptyset$. Let
\[
d_{max} = \max_{\ba \in S(t_1)}~d(\x,\ba).
\]
It is clear that $d_{max} < d(\x,\ba)$ for $\ba \in \hat{S}$. Then we have
\begin{equation*} 
\begin{array}{lrl}
 f(t_1) &   =  & \frac{1}{|S(t_1)|} \sum\limits_{\ba \in S(t_1)} d(\x,\ba) \\
        &   =  & \frac{|S(t_2)|}{|S(t_1)|} \times  \frac{1}{|S(t_2)|}  \sum\limits_{\ba \in S(t_1)} d(\x,\ba) \\
        &   =  & \frac{1}{|S(t_2)|} \Big[\frac{|\hat{S}|+|S(t_1)|}{|S(t_1)|} \times  \sum\limits_{\ba \in S(t_1)} d(\x,\ba) \Big]\\
        &   =  & \frac{1}{|S(t_2)|} \Big[\frac{|\hat{S}|}{|S(t_1)|} \sum\limits_{\ba \in S(t_1)} d(\x,\ba)+\sum\limits_{\ba \in S(t_1)} d(\x,\ba)\Big]\\
        & \leq & \frac{1}{|S(t_2)|} \Big[|\hat{S}| d_{max} + \sum\limits_{\ba \in S(t_1)} d(\x,\ba) \Big] \\
        & = & \frac{1}{|S(t_2)|} \Big[|\hat{S}| d_{max} + \sum\limits_{\ba \in S(t_2)} d(\x,\ba) -\sum\limits_{\ba \in \hat{S}} d(\x,\ba) \Big]\\
        &   <  & \frac{1}{|S(t_2)|} \sum\limits_{\ba \in S(t_2)} d(\x,\ba) \\
        &   =  & f(t_2).
\end{array}
\end{equation*}
The fact that $f$ is a step function follows from the finiteness of the set $A$. This completes the proof.
\end{proof}

\begin{figure*}[!h]
 \centering
    \subfloat[Liver Disorders]    {
    \includegraphics[width=0.4\textwidth]{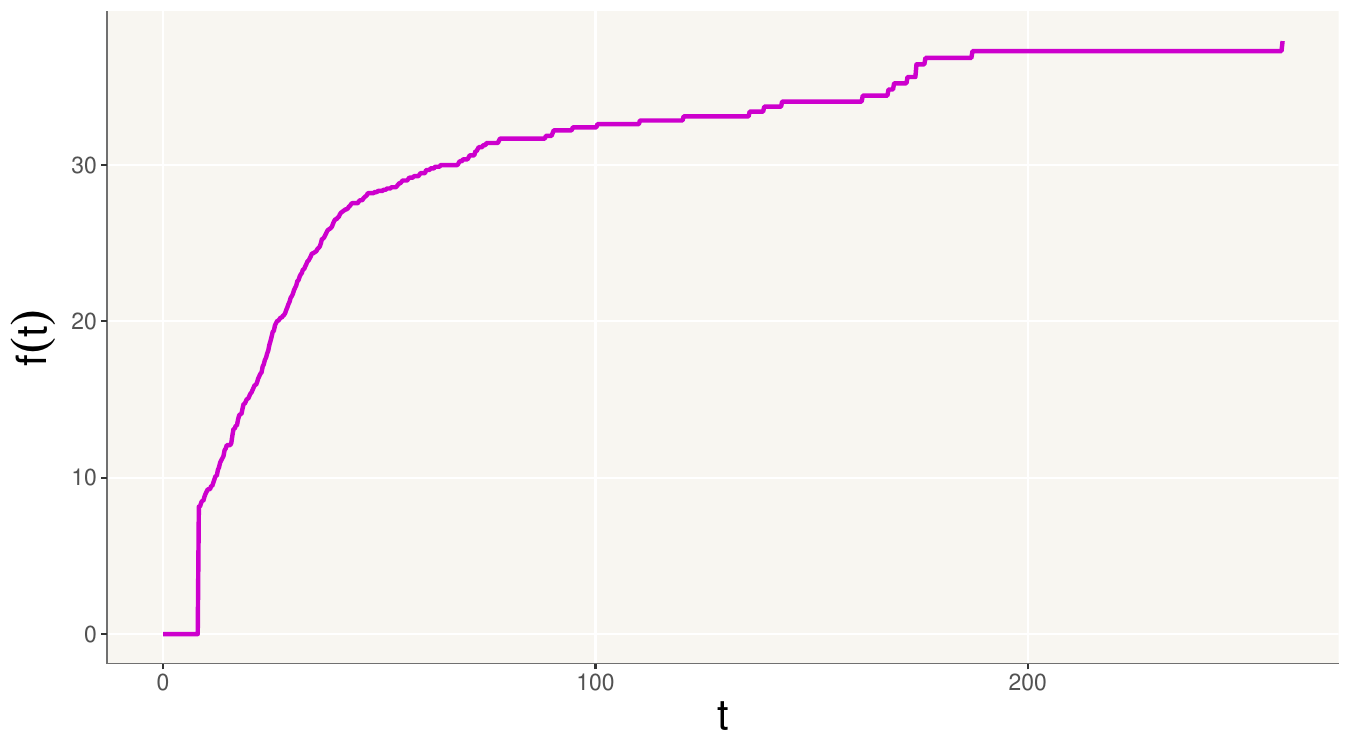}} \hspace{6pt}
    \subfloat[Ionosphere]{\includegraphics[width=0.4\textwidth]{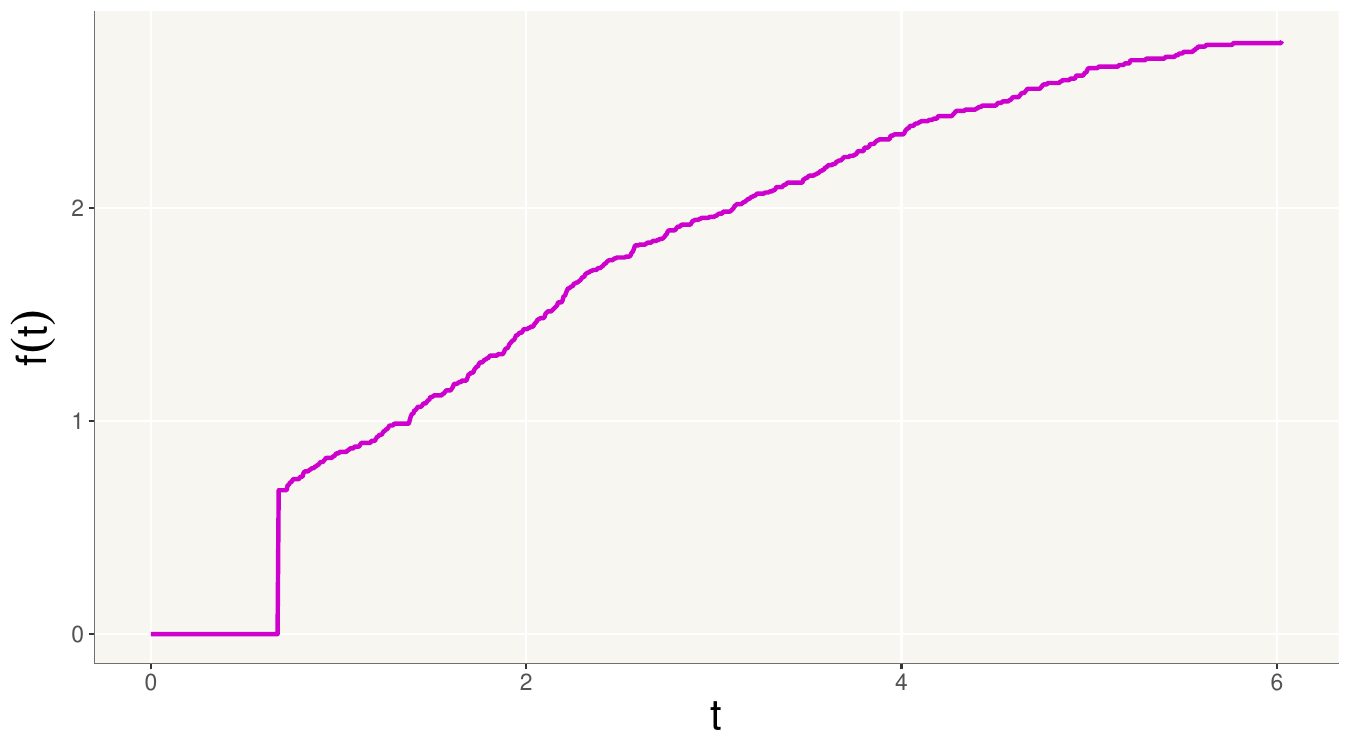}}
   \caption{Compactness functions}
 \label{CF-liverconstrained}
\end{figure*}

Using two real-world datasets from \cite{Duagraff2019} and the squared Euclidean distance as the similarity measure, we illustrate the compactness function, defined in \eqref{scatfunc}, in Figure \ref{CF-liverconstrained}. The dataset Liver Disorders contains 345 points with 6 attributes, while Ionosphere comprises 351 points with 34 attributes. The figure reveals several subintervals where the compactness function remains constant, indicating that no data points lie between the two concentric spheres corresponding to these intervals. These subintervals are longer for Liver Disorders compared to the Ionosphere. In both cases, the regions surrounding the centroids do not contain data points.

Compactness generally refers to the property of occupying minimal space. When applied to a dataset, it reflects how densely and uniformly data points are distributed: datasets with large empty regions exhibit low compactness. The compactness function is used to detect such sparse areas. In the graphs of compactness functions for datasets Liver Disorders and Ionosphere (Figure \ref{CF-liverconstrained}) these regions correspond to constant subintervals. The shorter these  subintervals are, the higher the compactness of the dataset.

Let $d_i= d(\ba_i,\x)$ for $\ba_i \in A,~i=1,\ldots,m,$ and define 
\begin{align*}
D= \Big\{d_1,d_2,\ldots,d_m  \Big\}.
\end{align*}
Sort these distances in ascending order and set $\hat  d_0=0$. Denote the ordered set by
\[
\hat{D} = \Big\{\hat{d}_0,\hat{d}_1,\hat{d}_2,\ldots,\hat{d}_p \Big\},
\]
where $\hat{d}_{i-1} < \hat{d}_i,~i=1,\ldots,p,~p \leq m.$ If $\x \in A$, then $\hat{d}_1=\hat{d}_0=0$; otherwise $\hat{d}_1>0$. Moreover, $\hat{d}_p = R_A$. Finally, if $\hat{d}_p > 0$, we normalize by dividing each element of $\hat{D}$ by $\hat{d}_p$ and construct the set
\[
{\cal D} = \Big\{\bar{d}_0,\bar{d}_1,\bar{d}_2,\ldots,\bar{d}_p \Big\},
\]
where $\bar{d}_i = \hat{d}_i/\hat{d}_p,~i=1,\ldots,p, ~\bar{d}_0=0$ and $\bar{d}_p = 1$. Moreover, $\bar{d}_{i-1} < \bar{d}_i,~i=2,\ldots,p$. We set ${\cal D}=\{0,0\}$ if $\bar{d}_p=0$.

Using the set ${\cal D}$, we define the intervals  $(\bar{d}_{i-1}, \bar{d}_i],~i=1,\ldots,p$. The compactness function $f$ remains constant within each open interval $(\bar{d}_{i-1}, \bar{d}_i),~i=1,\ldots,p$. These intervals vary in length: some are very short, whereas others are much longer (see Figure \ref{CF-liverconstrained} for datasets Liver Disorders and Ionosphere). For any given $\varepsilon > 0$, some intervals have $\bar{d}_i-\bar{d}_{i-1} \leq \varepsilon$, while others satisfy $\bar{d}_i-\bar{d}_{i-1} > \varepsilon$. Accordingly, the index set $P=\{1,\ldots,p\}$ can be partitioned into two disjoint subsets:
\begin{align*}
 Q_1(\varepsilon) = \Big\{i \in P: ~\bar{d}_i-\bar{d}_{i-1} \leq \varepsilon \Big\}, \quad \mbox{and} \quad Q_2(\varepsilon) = \Big\{i \in P: ~\bar{d}_i-\bar{d}_{i-1} > \varepsilon \Big\}.
\end{align*}
Note that as $\varepsilon$ increases, the cardinality of the set $Q_1(\varepsilon)$ increases, while the cardinality of  $Q_2(\varepsilon)$ decreases. We now define the following quantities:
\begin{align*}
\varepsilon_1 = \min_{i \in P} ~\big(\bar{d}_i-\bar{d}_{i-1}\big),\quad \mbox{and} \quad \varepsilon_2 = \max_{i \in P}~\big(\bar{d}_i-\bar{d}_{i-1}\big).
\end{align*}
For simplicity, from now on, we assume that $\bar{d}_1 > \bar{d}_0$. Then $\varepsilon_1 > 0$. If $\varepsilon<\varepsilon_1$, then $Q_1(\varepsilon) = \emptyset$ and $Q_2(\varepsilon) = P$. Conversely, if $\varepsilon \geq \varepsilon_2$, we have  $Q_1(\varepsilon) = P$ and $Q_2(\varepsilon) = \emptyset$. In particular, $Q_1(0)=\emptyset$ and $Q_2(0)=P$.

\begin{figure} [!h]
 \centering
 \includegraphics[width=0.35\textwidth]{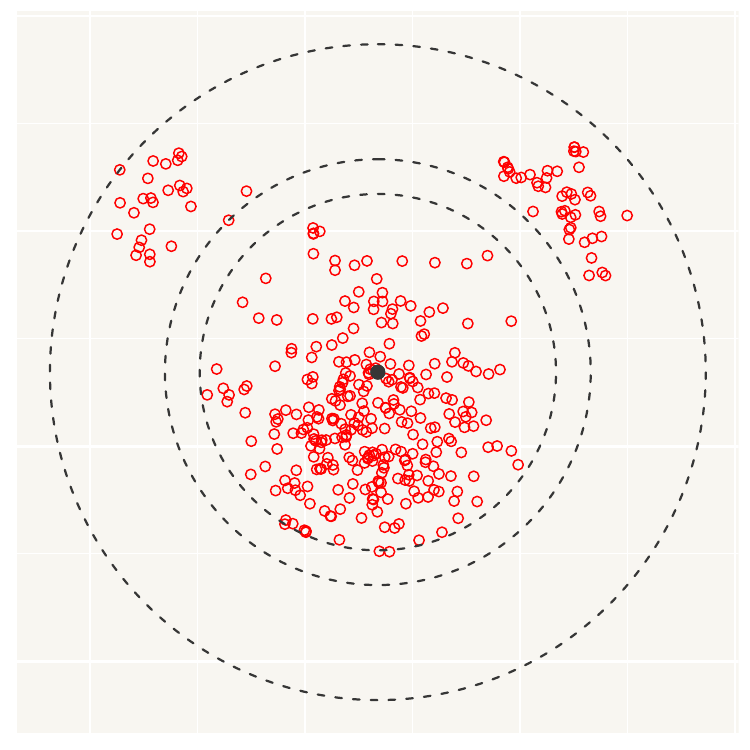} 
 \vspace{-5mm}
 \caption{Illustration of sets $V_1(\varepsilon),~V_2(\varepsilon) $ and $\bar{V}_1(\varepsilon)$.}
 \label{setsuv2}
\end{figure}

For a given $\varepsilon > 0$, define
\begin{align*}
U(\varepsilon) &= \bigcup_{i \in Q_1(\varepsilon)} \Big(\bar{d}_{i-1},\bar{d}_i\Big].
\end{align*}
Some of the intervals $(\bar{d}_{i-1},\bar{d}_i], ~i \in Q_1(\varepsilon)$ are adjacent. By merging these connected intervals into a single interval, we obtain a new collection of $\bar{p} \leq |Q_1(\varepsilon)|$ intervals. Let the endpoints of the merged intervals be denoted by $r_{j-1}$ and $r_j,~j=1,\ldots,\bar{p}$. Then, the set $U(\varepsilon)$ can be expressed as:
\begin{align*}
U(\varepsilon) &= \bigcup_{j=1}^{\bar{p}} \Big(r_{j-1},r_j\Big].
\end{align*}
Consider the set $V_j(\varepsilon) = \big\{\ba \in A: r_{j-1} < d(\x,\ba) \leq r_j \big\}, ~j= 1,\ldots,\bar{p}.$ The set $ \bar{V}_i(\varepsilon) = \big\{\ba \in A:~\bar{d}_{i-1}< d(\x,\ba) < \bar{d}_i \big\}= \emptyset$ for any $i \in Q_2(\varepsilon)$. The sets $V_j(\varepsilon),j=1,\ldots,\bar{p}$ and $\bar{V}_i(\varepsilon), i \in Q_2(\varepsilon)$ are illustrated in Figure \ref{setsuv2}. In this example,  $\bar{p}=2$ and $Q_2(\varepsilon) = \{1\}$: $V_1(\varepsilon)$ corresponds the inner ball, $V_2(\varepsilon)$ to the outer annulus, and $\bar{V}_1(\varepsilon)$ to the middle annulus.

\subsection{Compactness index} \label{compactness}
We define the compactness index using the compactness function $f$ introduced in \eqref{scatfunc}, the parameter $\varepsilon$, and the sets $V_j(\varepsilon),~j=1,\ldots,\bar{p},~\bar{V}_i(\varepsilon),~i \in Q_2(\varepsilon)$. Note that the sets $\bar{V}_i(\varepsilon),~i \in Q_2(\varepsilon)$ correspond to extended constant segments in the graphs of the compactness function $f$. 

It is important to assess how uniformly the data points are distributed within each non-empty set $V_j(\varepsilon),~j\in \{1,\ldots,\bar{p}\}$. However, this task becomes challenging as the number of attributes $n$ grows. To address this difficulty, we define a finite set of directions and examine the presence of data points along and around these directions.

Consider the positive spanning set $\mathds{E}=\big\{\bu_1,\ldots,\bu_{l}\big\}, ~l > n$ in the $n$-dimensional space $\R^n$, where $\|\bu_j\|=1,~j=1,\ldots,l$. Any vector in $\R^n$ can be expressed as a linear combination of these vectors with nonnegative coefficients. A commonly used positive spanning set is $\mathds{E}=\big\{\bu_1,\ldots,\bu_{2n}\big\},$ where $\bu_i=\e_i,~\bu_{n+i} = -\e_i,~i=1,\ldots,n$ with $\e_i$ denoting the $i$-th standard unit vector in $\R^n$.

Select a parameter $\eta \in (0,1]$, and using the set $V_j(\varepsilon)$, define the subset of $\mathds{E}$ as:
\begin{equation*} 
\bar{E}_j=\Big\{i \in \{1,\ldots,l\}: \max_{\ba \in V_j(\varepsilon), \ba \neq \x} \frac{\langle \ba - \x, \bu_i \rangle}{\|\ba-\x\|\|\bu_i\|} \geq \eta \Big\}, ~j= 1,\ldots,\bar{p}.
\end{equation*}
Here, $\langle \cdot, \cdot \rangle$ denotes the inner product on $\R^n$, $\x$ is the center of the set $A$, and $\|\cdot\|$ is the associated vector norm. The maximum ratio is equal to the cosine of the angle between the vectors $(\ba-\x)$ and $\bu_i$. By choosing $\eta \approx 0.7071,$ we ensure that this angle does not exceed $45^0$ for each $\bu_i \in \bar{E}_j$. If $|\bar{E}_j| = l$, the data points are nearly uniformly distributed within $V_j(\varepsilon)$; if $|\bar{E}_j|<l$, then some regions within $V_j(\varepsilon)$ are empty.

\begin{definition} \label{compcoef}
For a given $\varepsilon \geq 0,$ the number $\alpha_{j\varepsilon} = |\bar{E}_j|/l$ is called the $\varepsilon$-compactness coefficient of the set $V_j(\varepsilon),~j =1,\ldots,\bar{p}$.
\end{definition}

\begin{definition} \label{compnumber}
For a given $\varepsilon \geq 0,$ the number
\begin{align*} 
c_A(\varepsilon) = 1 - \Big[\sum_{j=1}^{\bar{p}} (1-\alpha_{j\varepsilon}) (r_{j} - r_{j-1}) +\sum_{i \in Q_2(\varepsilon)} \big(\bar{d}_{i}-\bar{d}_{i-1} -\varepsilon \big)\Big]
\end{align*}
is called the $\varepsilon$-compactness index of the set $A$.
\end{definition}

The compactness index $c_A(\varepsilon)$ has the following properties:
\begin{itemize}[nosep]
\item If $\varepsilon = 0$, then $Q_1(0)=\emptyset,~\bar{p}=0, ~Q_2(0)=\{1,\ldots,p\}$ and $c_A(0) = 1- \sum_{i \in Q_2(0)} (\bar{d}_i-\bar{d}_{i-1}) = 0;$
\item If $\varepsilon=1,$ then $\bar{p}=1, ~Q_1(\varepsilon)=\{1\}$ and $Q_2(\varepsilon)=\emptyset$. In this case, $V_1(1) = A,~r_1=1, r_0=0$. Let $\alpha_{11}$ be the compactness coefficient of the set $A$. Then $c_A(1) = 1- (1-\alpha_{11}) (r_1 - r_0) = \alpha_{11}.$
\end{itemize}

The parameter $\varepsilon$ is chosen to be moderate -- not too large and not too small. Since the values $\bar{d}_i,~i=1,\ldots,p$ are scaled within the interval $[0,1]$, $\varepsilon$ can typically be set to $0.01$ (or $0.001$) for small datasets and 0.0001 for large datasets.

\begin{figure}[h]
	\centering
    \includegraphics[trim=0 0 0 40,clip,width=0.5\textwidth]
    {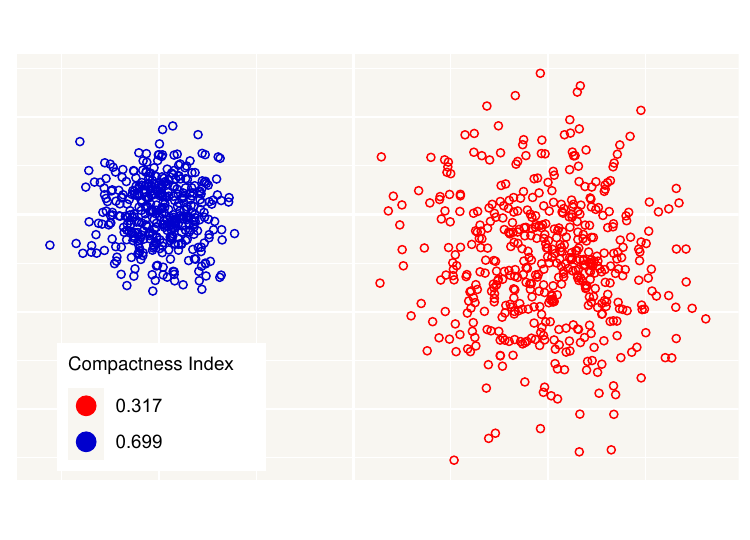}\vspace{-5mm}
	\caption{Compactness indices of clusters in the synthetic data.}
	\label{compTD2}
\end{figure}

Using a synthetic dataset with two clusters (Figure \ref{compTD2}), we illustrate the definition of the compactness index. The blue cluster is clearly more compact than the red cluster. Although the points in the red cluster are less densely packed, they are distributed in an almost uniform manner. For $\varepsilon=0.01$, the compactness index is \(0.699\) for the blue cluster and \(0.317\) for the red cluster. 

\begin{figure*}[h]
	\centering
	\subfloat[Liver Disorders]{
	\includegraphics[width=0.40\textwidth]{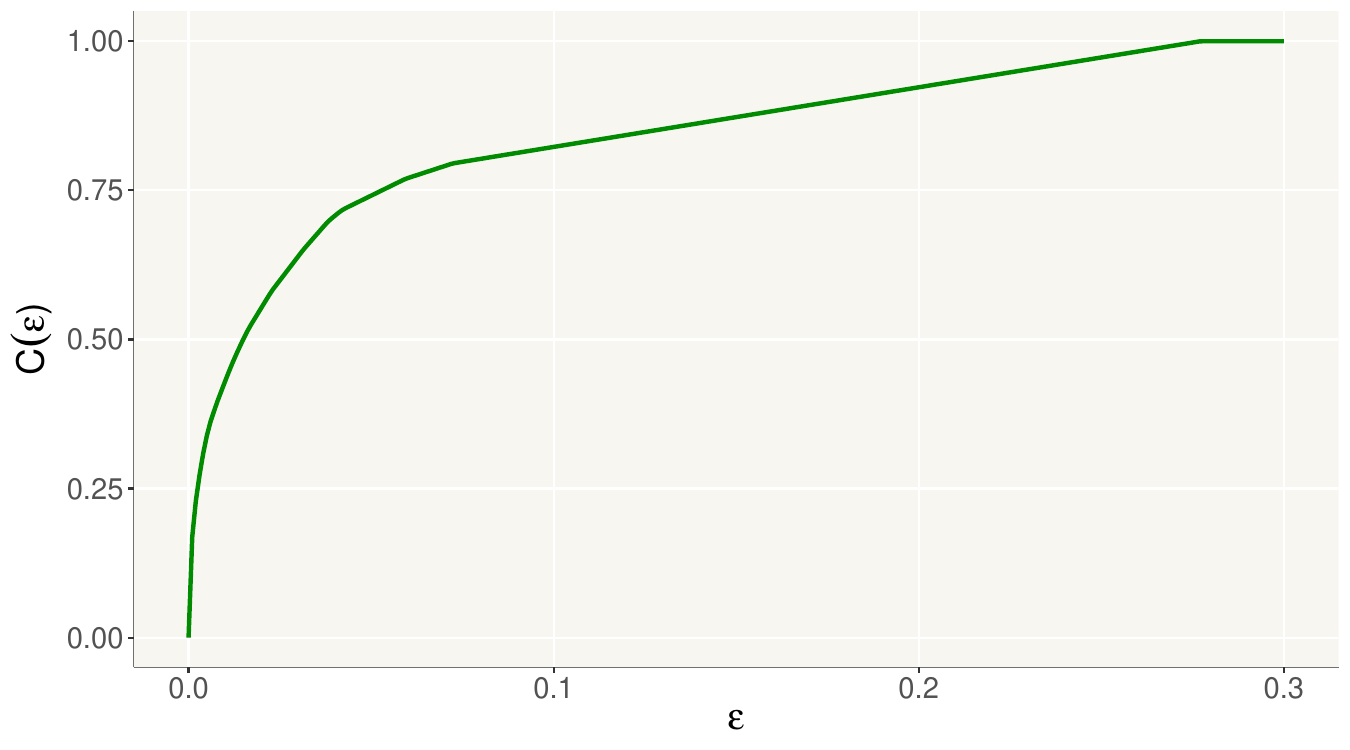}} \hspace{6pt}
	\subfloat[Ionosphere]{\includegraphics[width=0.40\textwidth]{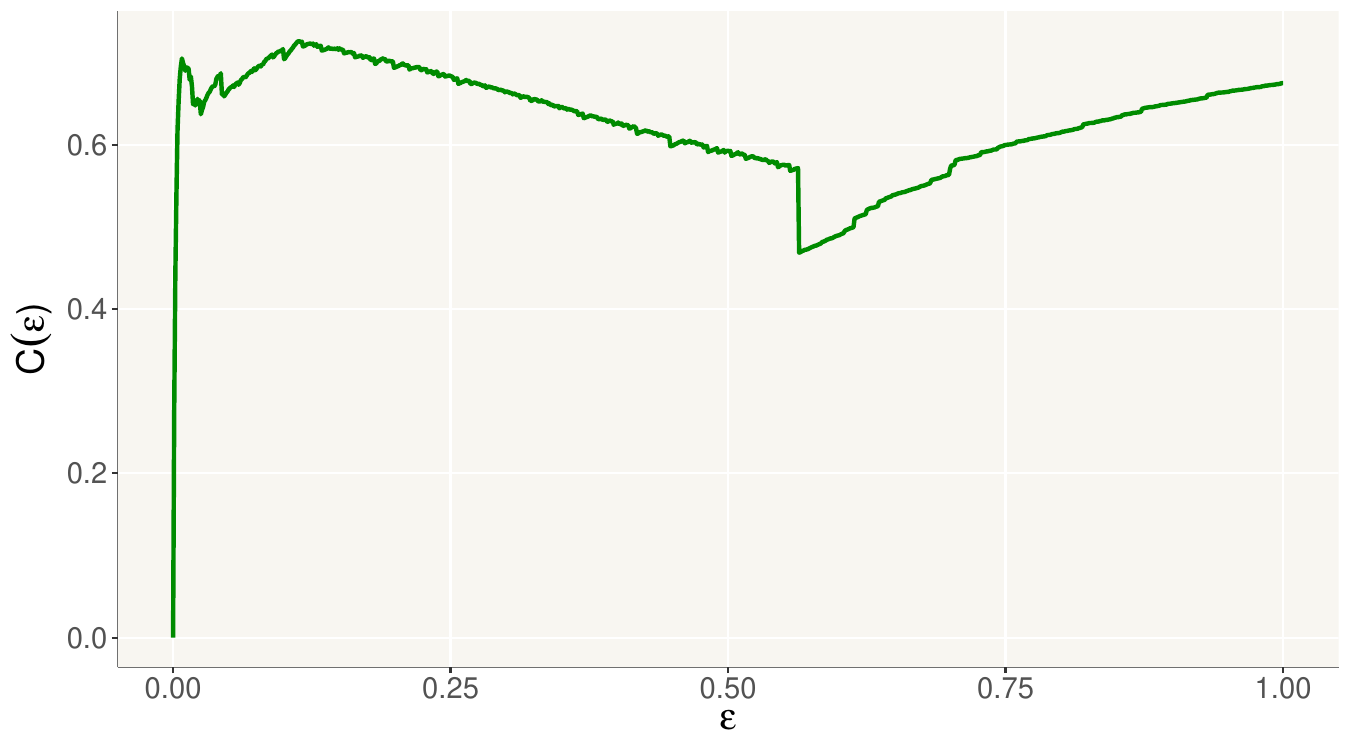}}
	\caption{Dependence of compactness index $c_A$ on $\varepsilon$.}
	\label{CF-liver-Ionos2}
\end{figure*}

Figure \ref{CF-liver-Ionos2} shows how the compactness index $c_A(\cdot)$ varies with $\varepsilon \in [0,1]$ for datasets Liver Disorders and Ionosphere. The plots indicate that the points in Liver Disorders are more evenly distributed overall, whereas in the Ionosphere most points are concentrated near the center.

\begin{definition} \label{compactdef2}
Let $\varepsilon > 0$ be given, $\bar{A} = \{A_1,\ldots,A_k\}, ~k> 1$ be a $k$-partition of the set $A$, and $c_{A_j}(\varepsilon)$ be an $\varepsilon$-compactness index of the cluster $A_j, j=1,\ldots,k$. The number
\begin{align}\label{epsiloncompactdis}
C_k(\varepsilon) = \frac{1}{|A|} \sum_{j=1}^k |A_j| c_{A_j}(\varepsilon)
\end{align}
is called the $\varepsilon$-compactness index of the cluster partition $\bar{A}$.
\end{definition}

\begin{figure}[h]
 \centering
   \includegraphics[trim=0 0 0 10,clip,width=0.6\textwidth]{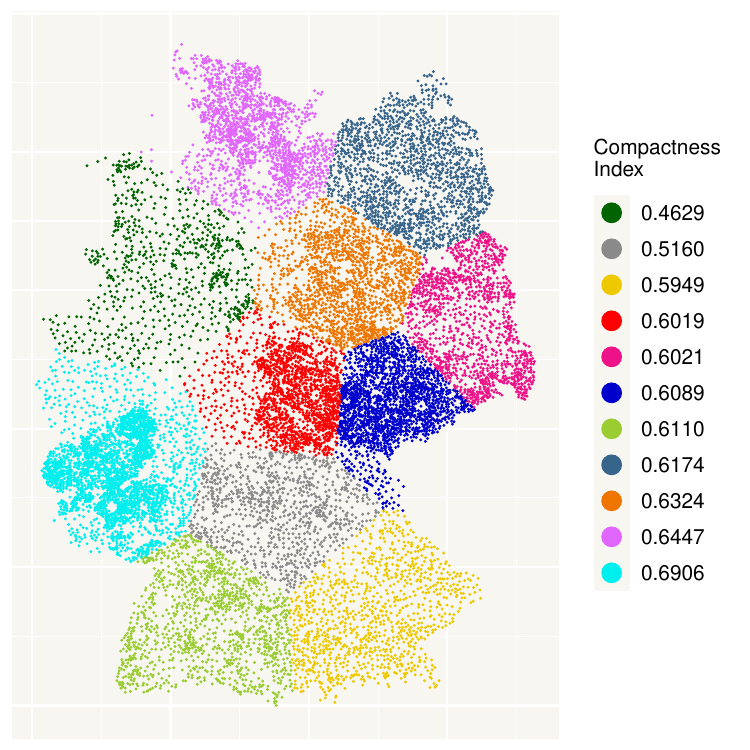}\vspace{-5mm}
   \caption{Compactness indices of clusters - D15112}
 \label{Comp-d15112}
\end{figure}

\begin{figure}[!t]
 \centering
   \includegraphics[trim=0 0 0 0,clip,width=0.6\textwidth]{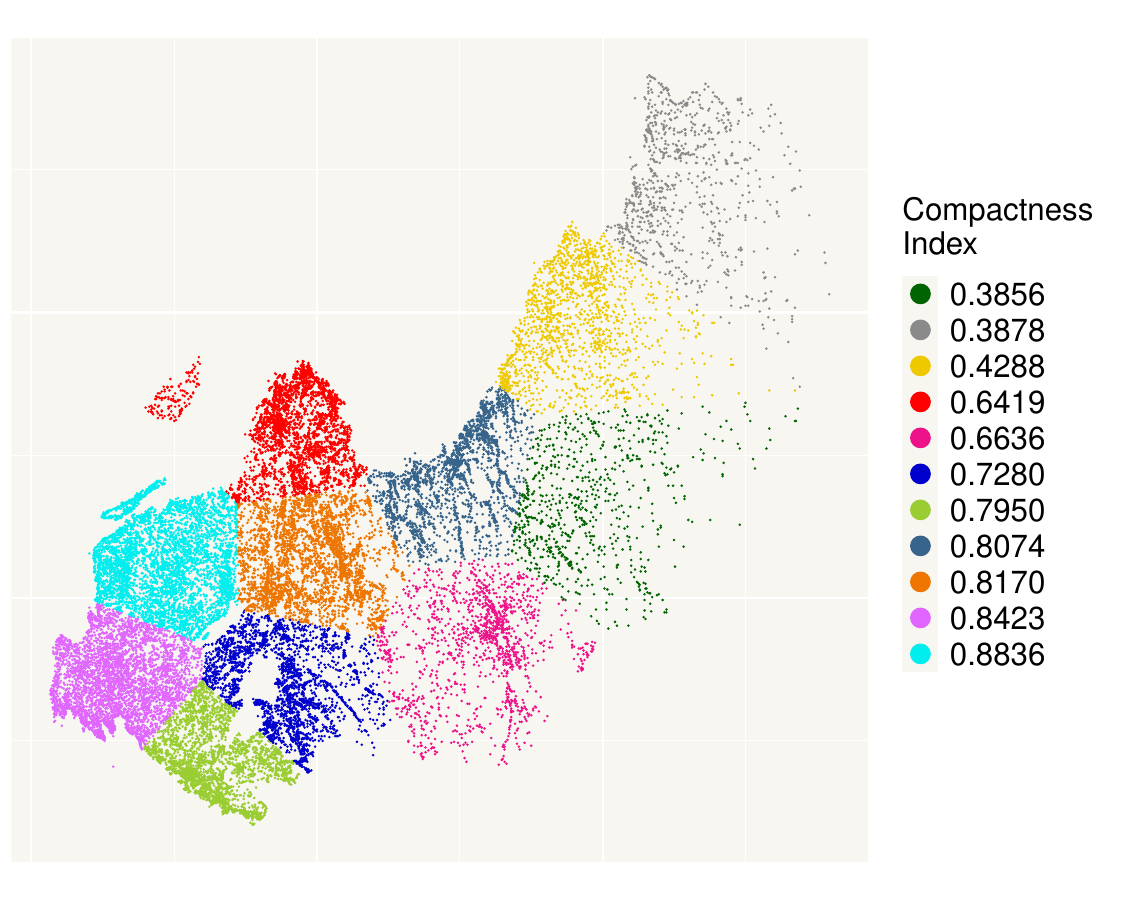}\vspace{-5mm}
   \caption{Compactness indices of clusters - SW24978}
 \label{Comp-SW24978}
\end{figure}

As illustrative examples, we consider two real-world datasets: D15112 and SW24978, each containing two attributes. D15112 consists of 15,112 points, while SW24978 contains 24,978 points. Using $\varepsilon = 0.001$, we obtain compactness indices $C_1(0.001)=0.973$ for D15112 and $C_1(0.001)=0.954$ for SW24978, indicating that D15112 is more compact than SW24978. We then partition each dataset into $11$ clusters and again set $\varepsilon=0.001$. The resulting clusters and their corresponding compactness values are shown in Figures~\ref{Comp-d15112} and \ref{Comp-SW24978}. Cluster labels are arbitrary and ordered according to the compactness value. For these partitions, $C_{11}(0.001) = 0.615$ for D15112 and $C_{11}(0.001)=0.743$ for SW24978, showing that, at the cluster level, SW24978 is more compact than D15112. This is due to the presence of several highly compact clusters in SW24978, while its less compact clusters contain only a small fraction of the total points. These findings demonstrate that the compactness of an entire dataset can differ significantly from the compactness of its cluster partition.

Next, we analyze the computational complexity of the compactness index in terms of both time and space requirements. Recall that the set $A$ contains $m$ points described by $n$ attributes, and assume that the similarity measure is derived from a distance function. The computation of the compactness index consists of the following stages:
\begin{enumerate} [nosep]
\item 
Determining the center of the data set requires $O(mn)$ operations;
\item  
Calculating the distances between all data points and the center, followed by normalization, requires $O(mn)$ operations;
\item 
Ordering the distances in ascending order requires $O(m \log m)$ operations;
\item  
Given $\varepsilon > 0$, dividing the intervals into two groups requires $O(m)$ operations;
\item 
Determining empty and non-empty annuli requires $O(m)$ operations;
\item 
Computing the compactness coefficient for each annulus requires $O(mn)$ operations;
\item 
The final compactness index is obtained in $O(m)$ operations.    
\end{enumerate}
Combining the costs of all stages and assuming that $n=O(\log m)$, the overall time complexity is dominated by $O(mn)$. The space complexity is also $O(mn)$. Therefore, both the time and space complexities of the compactness index are estimated as $O(mn)$.

\vspace{-3mm}

\section{Adjacent sets and separability of clusters} \label{separable}
For any two finite point sets in the $n$-dimensional space $\R^n,$ the boundary separating them consists of points that lie adjacent to both sets. This collection of boundary points determines the separability of these sets. To identify such points, we introduce the notion of adjacent sets between two sets.

\subsection{Adjacent sets}
\begin{figure}[h]
\centering
\subfloat[Synthetic dataset]{
\includegraphics[trim=40 10 50 20,clip,width=0.3\textwidth, angle=90]{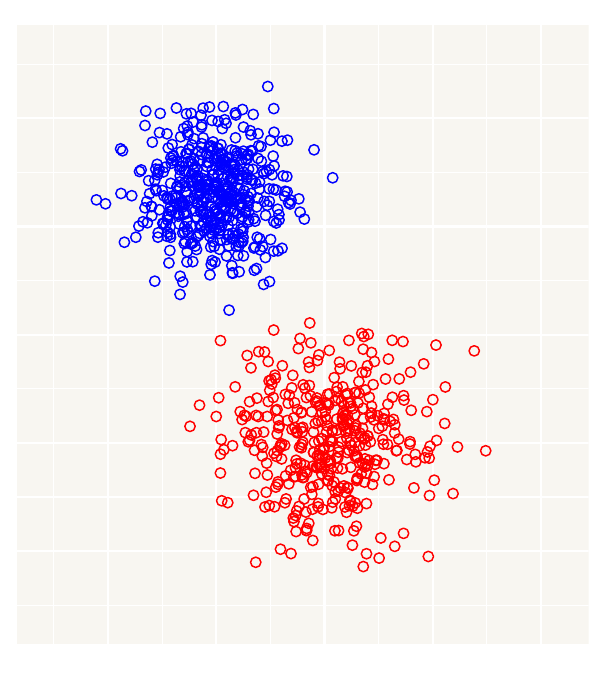}}
\subfloat[Set $\bar{Z}_{12}$ (green points)]{
\includegraphics[trim=40 10 50 20,clip,width=0.3\textwidth, angle=90]{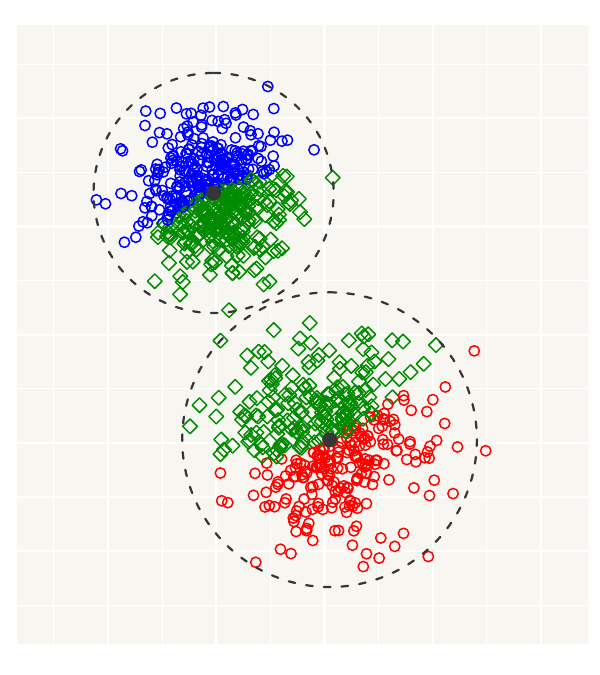}}
\caption{Illustration of an adjacent set using a synthetic data.}
\label{Q1Q2}
\end{figure}

Consider two finite point sets $\mathcal{A}_1$ and $\mathcal{A}_2$ in the $n$-dimensional space $\R^n$. Let $\x_1$ and $\x_2$ be the centers of these sets, respectively. Denote the distance between these centers by  $d_{12} = d(\x_1,\x_2).$
\begin{definition} \label{adjacent} The set
\begin{equation} \label{adset01}
Z_{12} = \Big\{\ba \in \mathcal{A}_1:~d(\x_2,\ba) \leq d_{12} \Big\}
\end{equation}
is called the subset of $\mathcal{A}_1$ adjacent to $\mathcal{A}_2$ and \begin{equation} \label{adset02}
Z_{21} = \Big\{\bb \in \mathcal{A}_2:~d(\x_1,\bb) \leq d_{12} \Big\}
\end{equation}
is called the subset of $\mathcal{A}_2$ adjacent to the set $\mathcal{A}_1$. The set $\bar{Z}_{12} = Z_{12}\cup Z_{21}$ is called the adjacent set between the sets $\mathcal{A}_1$ and $\mathcal{A}_2$.
\end{definition}
It follows from the definition that:
\begin{itemize}[nosep]
\item[1.] The set $Z_{12}$ contains all points in $\mathcal{A}_1$ whose distance to the center of $\mathcal{A}_2$ is less than or equal to the distance between the centers of $\mathcal{A}_1$ and $\mathcal{A}_2$;
\item[2.] Similarly, the set $Z_{21}$ consists of all points in $\mathcal{A}_2$ whose distance to the center of $\mathcal{A}_1$ is less than or equal to the distance between the centers of $\mathcal{A}_1$ and $\mathcal{A}_2$.
\end{itemize}
To illustrate the adjacent sets, we consider the synthetic dataset in Figure \ref{Q1Q2}(a). This set contains two clusters: red and blue. The adjacent set of these two sets is shown in Figure \ref{Q1Q2}(b), where the green points represent the set $\bar{Z}_{12}$.

\vspace{-3mm}

\begin{definition} \label{separable2}
Compute
\begin{equation*} 
\Delta_{12} = \max_{\ba \in Z_{12}} d(\x_1,\ba),\quad\mbox{and}\quad\Delta_{21} = \max_{\bb \in Z_{21}} d(\x_2,\bb).
\end{equation*}
We have the following:
\begin{itemize}[nosep]
 \item[1.] The number $\hat{\beta}_{12} =  d_{12} - \Delta_{12} - \Delta_{21}$ is called the margin between sets $\mathcal{A}_1$ and $\mathcal{A}_2;$
 \item[2.] The number $\bar{\beta}_{12} = \hat{\beta}_{12}/d_{12}$ is the scaled margin between sets $\mathcal{A}_1$ and $\mathcal{A}_2;$
 \item[3.] The number $\beta_{12} = 0.5(\bar{\beta}_{12}+1)$ is called the separability index for $\mathcal{A}_1$ and $\mathcal{A}_2$.
\end{itemize}
\end{definition}

\begin{proposition} \label{betabar}
	Let $\mathcal{A}_1, \mathcal{A}_2 \subset \mathbb{R}^n$ be any nonempty finite point sets and $\x_1$ and $\x_2$ be their centers, respectively. Assume that the following conditions are satisfied:
	\begin{align}
	d(\x_1,\ba)<d(\x_2,\ba),~\forall \ba \in \mathcal{A}_1,~\mbox{and}~d(\x_2,\bb)<d(\x_1,\bb),~\forall \bb \in \mathcal{A}_2.
    \label{as:prop}
	\end{align}
	Then $\beta_{12} \in [0,1]$.
\end{proposition}
\begin{proof} From the definition of sets $Z_{12}$ and $Z_{21}$, along with the assumption \eqref{as:prop} it follows that $d(\x_1,\ba) < d(\x_2,\ba) \leq d_{12}$ for all $\ba \in Z_{12}$ and $d(\x_2,\bb) < d(\x_1,\bb) \leq d_{12}$ for all $\bb \in Z_{21}$. Given that $\mathcal{A}_1$ and $\mathcal{A}_2$ are finite point sets, we have $\Delta_{12} \leq d_{12}$ and $\Delta_{21} \leq d_{12}$. This means $\hat{\beta}_{12} = d_{12} - \Delta_{12} - \Delta_{21} \geq -d_{12},$ which implies that $\bar{\beta}_{12} \geq -1$. Since $\Delta_{12} \geq 0$ and $\Delta_{21} \geq 0$, it follows that $\bar{\beta}_{12} \leq 1$. The value $\bar{\beta}_{12} = 1$ occurs when both $\mathcal{A}_1$ and $\mathcal{A}_2$ are singletons. In this case, we have $\Delta_{12} = \Delta_{21} = 0$, and therefore $\beta_{12} \in [0,1]$.
\end{proof}

Since assumption  \eqref{as:prop} holds for all clusters in any cluster configuration, Proposition \ref{betabar} applies to every pair of clusters. If $\beta_{12} > 0.5$, then the sets $\mathcal{A}_1$ and $\mathcal{A}_2$ are separable; when $\beta_{12} \leq 0.5$ they are inseparable. In Figure \ref{Q1Q2}(b), the two sets are separable, with a separability index of $\beta_{12}=0.524$.

Consider the set $A$ and its cluster distribution $\bar{A} = \{A_1,\ldots,A_k\}$. Let $\beta_{ij}$ be the separability index of the clusters $A_i$ and $A_j, ~i,j=1,\ldots,k,~i \neq j $. The matrix
\vspace{-2mm}
\[M = \left[
\begin{array}{cccc}
   0                  & \beta_{12} &  \cdots & \beta_{1k} \\
   \beta_{21} &        0           & \cdots  & \beta_{2k} \\
   \cdots          &   \cdots          & \cdots  &    \cdots       \\
   \beta_{k1} &  \beta_{k2} & \cdots  &    0              \\
\end{array}\right].
\]
is called the \textit{separability matrix} of the cluster distribution $\bar{A}$. This matrix is symmetric, that is, $\beta_{ij}= \beta_{ji},~i,j=1,\ldots,k$.

\begin{figure*}[!h]
\centering
\subfloat[DA1]{
\includegraphics[trim=0 0 0 10,clip,width=0.28\textwidth]{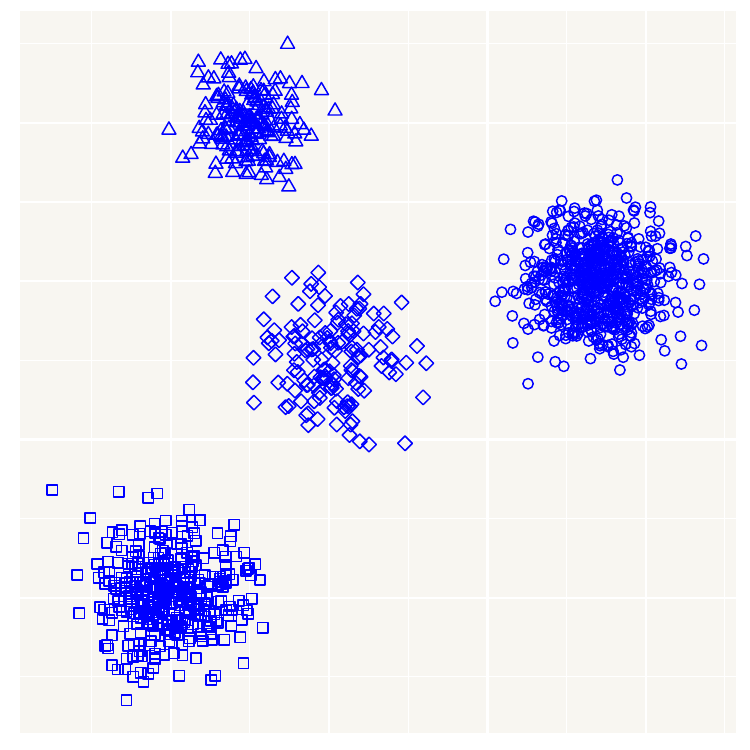}}
\subfloat[DA2]{
\includegraphics[trim=0 0 0 10,clip,width=0.28\textwidth]{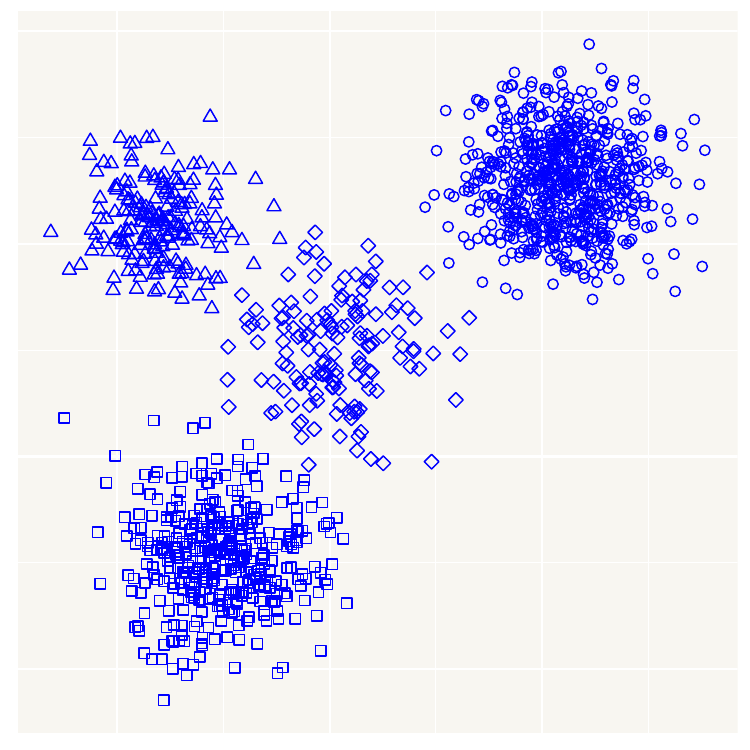}}
\subfloat[DA3]{
\includegraphics[trim=0 0 0 10,clip,width=0.28\textwidth]{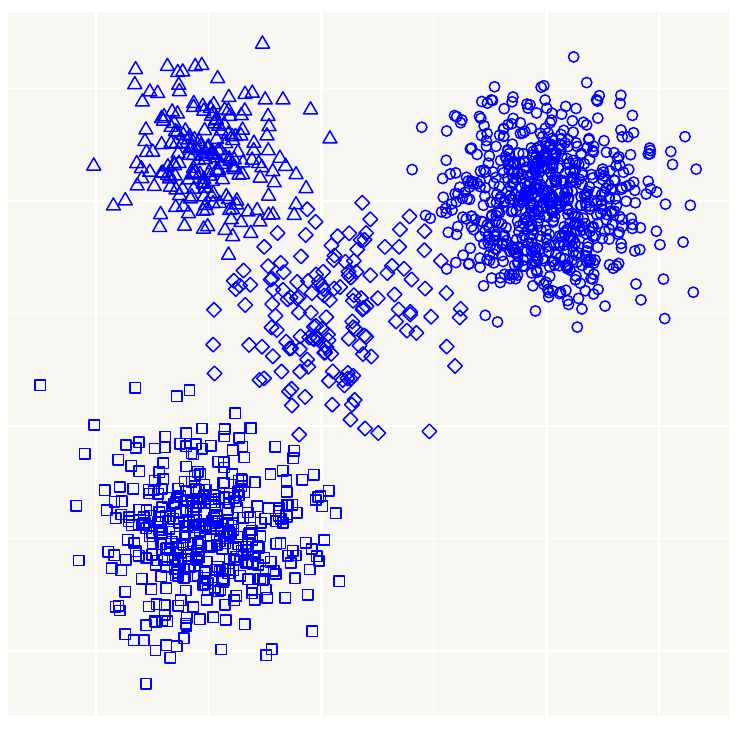}}
\caption{Synthetic datasets with 4 clusters and different separability}
\label{marmatrix}
\end{figure*}

Consider the synthetic datasets illustrated in Figure \ref{marmatrix}(a-c). Each dataset comprises four clusters, with data points represented by the symbols $\circo, \squareo,\triangleo$ and $\diamondo$. In the first dataset (DA1), the clusters are well separated; in the second (DA2), the clusters are closer but still clearly distinguishable; in the third (DA3), the clusters are intermixed. For each dataset, we compute the corresponding separability matrix $M$.

The separability matrix for the DA1 dataset is given below. We can see that all elements of $M$ are strictly greater than $0.5$, indicating that the clusters are well separated.
\vspace{-7mm}
\begin{align*}
M =
\begin{blockarray}{ccccc}
             &$\circo$& $\squareo$ & $\triangleo$ & $\diamondo$ \\
 \begin{block}{c[cccc]}
$\circo$     &  0     &    0.777   &     0.726    &    0.574    \\
$\squareo$   &  0.777 &      0     &     0.753    &    0.608    \\
$\triangleo$ &  0.726 &    0.753   &     0        &    0.645    \\
$\diamondo$  &  0.574 &    0.608   &     0.645    &       0     \\
\end{block}
\end{blockarray}.
\end{align*}

The separability matrix for DA2, given below, shows that the entries in the last row (or column) indicate that the $\diamondo$ cluster is not separated from the $\squareo$ and $\triangleo$ clusters. 
\vspace{-1mm}
\begin{align*}
M =
\begin{blockarray}{ccccc}
                  &$\circo$& $\squareo$&$\triangleo$& $\diamondo$ \\
 \begin{block}{c[cccc]}
$\circo$          &  0     &    0.712  &    0.677   &     0.509   \\
$\squareo$        &  0.712 &      0    &    0.513   &     0.372   \\
$\triangleo$      &  0.677 &    0.513  &      0     &     0.440   \\
$\diamondo$       &  0.509 &    0.372  &    0.440   &       0     \\
\end{block}
\end{blockarray}.
\end{align*}

The separability matrix for DA3, shown below, reveals that the central cluster ($\diamondo$) cannot be separated from any of the other clusters. Across all three datasets, these results demonstrate that the separability index defined in Definition \ref{separable2} effectively captures the degree of separability of the cluster.
\vspace{-2mm}
\begin{align*}
M =
\begin{blockarray}{ccccc}
                  &$\circo$& $\squareo$&$\triangleo$&$\diamondo$ \\
 \begin{block}{c[cccc]}
$\circo$          &  0     &    0.678  &    0.604   &    0.422   \\
$\squareo$        &  0.678 &      0    &    0.563   &    0.383   \\
$\triangleo$      &  0.604 &    0.563  &      0     &    0.383   \\
$\diamondo$       &  0.422 &    0.383  &    0.383   &      0     \\
\end{block}
\end{blockarray}.
\end{align*}

Next, we introduce the separability index for cluster distributions.
\begin{definition} \label{separable5}
Let $\bar{A} = \{A_1,\ldots,A_k\}$ be a $k$-partition of the set $A$ and $\beta_{ij}$ be the separability index between $A_i$ and $A_j$,~$i,j=1,\ldots,k, ~j \neq i$. For each $i=1,\ldots,k$, define
\[
\bar{s}_{ik} = \min_{j =1,\ldots,k, j \neq i} \beta_{ij},~i=1,\ldots,k.
\]
The number 
\begin{equation} \label{sepindextotal1}
s_k = \frac{1}{|A|} \sum_{i=1}^k |A_i| \bar{s}_{ik}
\end{equation}
is called the separability index of the cluster distribution $\bar{A}$. It is clear that $s_k \in [0,1]$.
\end{definition}

\begin{figure*}[!h]
\centering
\subfloat[]{
\includegraphics[trim=0 0 0 25,clip,width=0.23\textwidth]{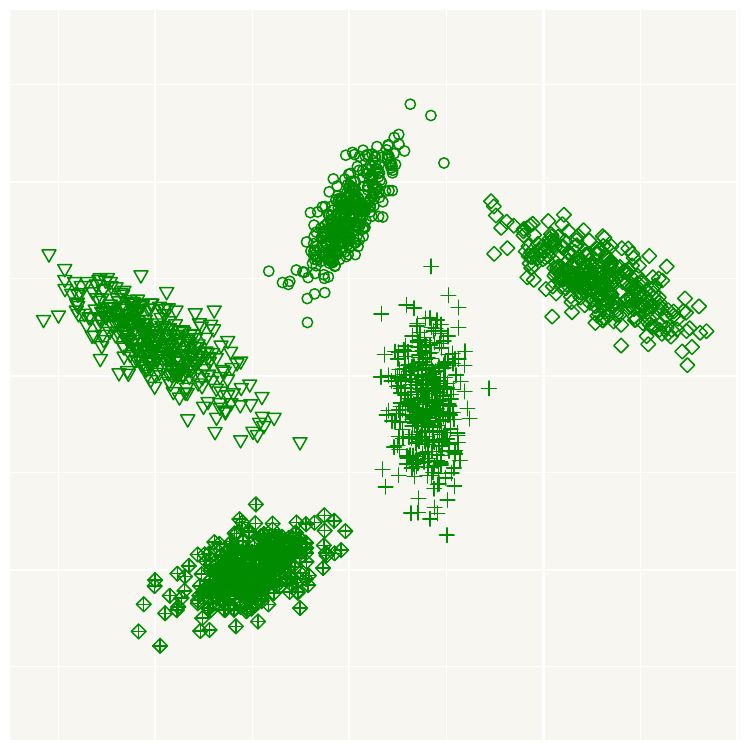}}
\subfloat[]{
\includegraphics[trim=0 0 0 25,clip,width=0.23\textwidth]{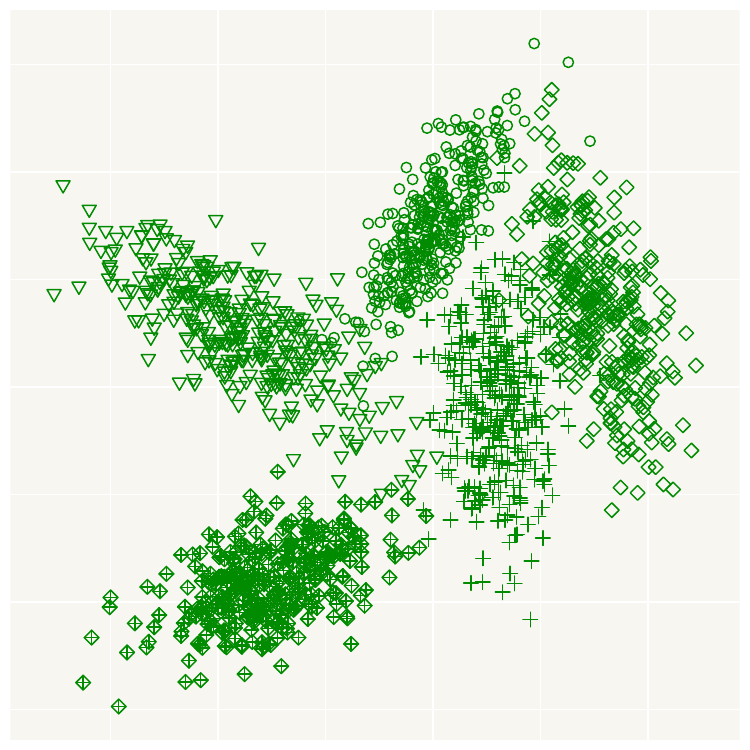}}
\subfloat[]{
\includegraphics[trim=0 0 0 25,clip,width=0.23\textwidth]{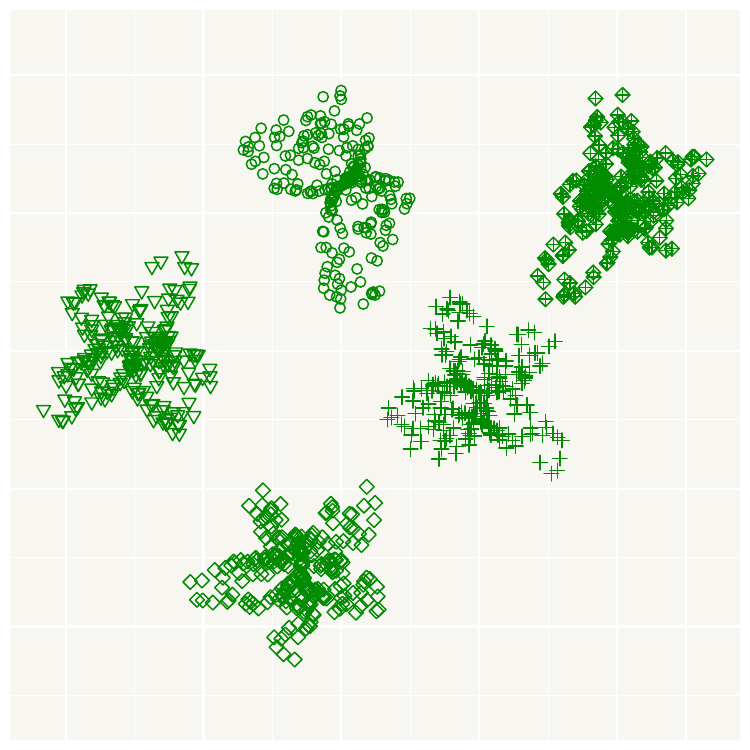}}
\subfloat[]{
\includegraphics[trim=0 0 0 25,clip,width=0.23\textwidth]{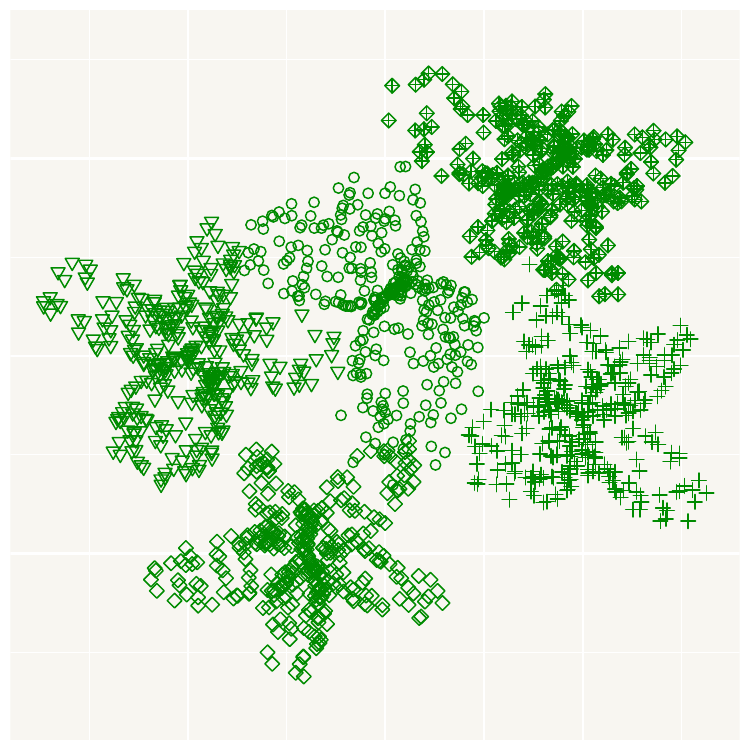}}
\caption{Synthetic datasets with elongated and irregular clusters}
\label{ellirr}
\end{figure*}

We study the separability index for a similarity measure based on the squared Euclidean distance, which implicitly assumes good performance when clusters are approximately ball-shaped. Although other similarity measures can be used, the separability index may, in principle, depend on the chosen measure and on cluster geometry. Nevertheless, the examples below suggest that this dependence is often limited. To illustrate this, we consider four synthetic datasets in Figure \ref{ellirr}, each containing elongated or nonconvex clusters. Datasets (a) and (b) have elongated clusters, with (b) obtained from (a) by reducing the inter-cluster distances. Datasets (c) and (d) contain nonconvex clusters, with (d) derived from (c) by moving the clusters closer together. The separability index values for different numbers of clusters are reported in Table \ref{tab:ellirrsynth}; specifically, $s_k(\bar{a})$ denotes the separability index computed for dataset $\bar{a}$ shown in Figure \ref{ellirr}, where $\bar{a} \in \{(a), (b), (c), (d)\}$. In all four cases, the separability index correctly identifies the true number of clusters, highlighting robustness across diverse cluster shapes.

\begin{table}[!htbp]
\centering\footnotesize
 \renewcommand{\arraystretch}{0.75} 
\begin{tabular}{lrrrr}
\toprule
$k$ &$s_k(a)$&$s_k(b)$& $s_k(c)$ & $s_k(d)$ \\
\cmidrule(lr){2-5}
2 & 0.245 & 0.052 & 0.292 & 0.228   \\
3 & 0.178 & 0.292 & 0.331 & 0.272   \\
4 & 0.343 & 0.205 & 0.414 & 0.217   \\
5 & {\bf 0.441} & {\bf 0.293} & {\bf 0.545} & {\bf 0.304}   \\
6 & 0.427 & 0.227 & 0.497 & 0.249   \\
7 & 0.406 & 0.275 & 0.426 & 0.208   \\
\bottomrule
\end{tabular}
\caption{Values of separability index for different number of clusters.}
\label{tab:ellirrsynth}
\end{table}

\vspace{5mm}


We now discuss the time and space complexity of the separability index. Assume that a $k$-partition of the dataset $A$ is given, where $A$ contains $m$ data points with $n$ attributes. The computation of the separability index is based on the distance matrix between data points and cluster centers. This matrix has dimensions $m \times k$ and therefore requires evaluating $mk$ distances. Since computing each distance involves $n$ attributes, the overall cost of constructing the distance matrix is $O(mnk)$. Additional operations required for computing the separability index are of lower order and do not affect the overall complexity. Accordingly, the time complexity of the separability index is estimated as $O(mnk)$. With respect to memory requirements, the computation requires storing the dataset and the $k$ cluster centers. Using these inputs, the distance matrix between data points and cluster centers is constructed, together with the cluster assignment information for all data points. As a result, the overall space complexity is dominated by the storage needed for the dataset and can be estimated as $O(mn)$.

\vspace{-3mm}
\section{Determining the number of clusters} \label{optimalnum}
Deciding on the ``true" number of clusters can be guided by various criteria. A common strategy is to select the number of clusters that produces a distribution with maximum compactness and maximum separation. In this sense, determining the ``true" number of clusters becomes a multiobjective problem, where the chosen number represents a balance between compactness and separability.

Both compactness and separability indices are scaled in the range $[0,1]$, allowing a direct comparison between them. Each cluster solution is represented as a point in a two-dimensional space, where the first coordinate corresponds to the compactness index and the second to the separability index. All such points are within the box $[0,1] \times [0,1] \subset \mathbb{R}^2$, and the set of these points is called the \textit{decision-space plot}. 

Within this plot, the solution corresponding to the optimal number of clusters is found among the non‑dominated points. It is worthwhile to choose the solution with the highest separability index among the non-dominated solutions to represent the optimal number of clusters as it provides the most distinct cluster structure. The number of clusters associated with other non‑dominated points may also be considered plausible candidates for the ``true" number of clusters.

To scalarize this two-objective problem, we proceed as follows. For a given $k$-partition of the dataset and any $\varepsilon \in [0,1]$, we combine the compactness and separability indices into a single criterion:
\begin{equation} \label{combined}
T_k(\varepsilon) = \frac{1-C_k(\varepsilon)}{s_k},
\end{equation}
where $C_k(\varepsilon)$ and $s_k$ are defined in \eqref{epsiloncompactdis} and \eqref{sepindextotal1}, respectively. Among all the non-dominated points in the decision-space plot, the minimal value of $T_k(\varepsilon)$ identifies the true number of clusters. This scalarization of the two objectives yields a selection rule that favors cluster configurations with higher separability.

Alternative approaches for scalarization - such as using a weighted sum or the product of these two indices - can also be applied. However, the weighted-sum approach requires careful selection of the weights, and an inappropriate choice may lead to suboptimal results. In contrast, a scalarization scheme based on the product of the two indices generally produces outcomes similar to those obtained using the ratio-based approach adopted in this paper.

\vspace{-3mm}

\section{Numerical experiments} \label{numerical}
In this section, we assess the effectiveness of the proposed indices for estimating compactness, separability, and the number of clusters, using synthetic and real‑world datasets. We also compare the new combined compactness–separability index with several established cluster validity indices, whose definitions are provided in the corresponding references:
\begin{itemize}[nosep]
\item [1.] The average silhouette index ($S_{av}$) \cite{Rousseeuw1987};
\item [2.] Davies-Bouldin ($DB$) index \cite{Bouldin1979};
\item [3.] Calinski-Harabasz ($CH$) index \cite{Calinski1974};
\item [4.] Dunn ($Dn$) index\cite{Dunn1974};
\item [5.]Xie-Beni ($XB$) index \cite{Xiebeni1991};
\item [6.]Absolute $G$-indices ($G_{str}$ and $G_{rex}$) \cite{Zimek2020}.
\end{itemize}
Note that the index $S_{av}$ represents the average silhouette value at all data points. Clustering was performed using the discrete gradient-based incremental clustering algorithm, which builds clusters sequentially, beginning with a single cluster. At each iteration, the results of the previous step are used to initialize the cluster centers for the next. Details of this algorithm and its implementation can be found in \cite{bagkartah2024}, and its source code is available in \cite{Github}. The compactness index is defined in equation \eqref{epsiloncompactdis}, the separability index is computed using equation \eqref{sepindextotal1}, and the combined index is given by equation \eqref{combined}. The source code of the proposed indices is available in \cite{Github}.

\subsection{Synthetic datasets}
We consider the following synthetic datasets:
\begin{itemize}[nosep]
\item [1.] Datasets A1, A2, A3, Unbalance, and Dim256 from \cite{Franti2018};
\item [2.] Datasets DA1, DA2, and DA3, generated using a mixture of statistical distributions, available at  \cite{DataSets} and illustrated in Figure \ref{marmatrix}.
\end{itemize} 
The A1, A2, and A3 datasets contain numerous compact circular clusters, with the number of observations and clusters increasing across the three sets: A1 has $3,000$ observations and $20$ clusters, A2 has $5,250$ observations and $35$ clusters, and A3 has $7,500$ observations and $50$ clusters. Each cluster contains $150$ points.

The Unbalance dataset contains $6,500$ observations that form eight Gaussian clusters. Among these, three clusters are dense with $2,000$ observations each, while the remaining five clusters each contain $100$ observations. Dim256 consists of $1,024$ observations in $16$ well‑separated Gaussian clusters, each described by $256$ features.

The second group includes two‑dimensional datasets with four clusters and $1,450$ observations, as illustrated in Figure \ref{marmatrix}.  

\subsection{Real-world datasets}
In addition to the Liver Disorders, Ionosphere, D15112, and SW24978 datasets we also employ the following real-world datasets from \cite{Duagraff2019}: TSPLIB3038 ($m=3,038; n = 2$), Land Satellite ($m=6,435; n = 36$), Shuttle Control ($m=58,000; n = 9$), and Localization Data for Person Activity ($m=164,860; n = 3$). 

\subsection{Results}
The results are summarized in Tables \ref{tab:a1}-\ref{tab:PersonActivity}, where $k$ represents the number of clusters and $T_k$ is the combined compactness-separability index defined in \eqref{combined}. The best results correspond to the highest values of the indices $S_{av}$, $Dn$, $CH$, $G_{str}$, and $G_{rex}$, and the lowest values of the indices $T_k$, $DB$, and $XB$. These values are highlighted in bold within the tables. Table \ref{tab:a1} reports the results for the synthetic dataset A1, which contains $20$ clusters. All cluster validity indices consistently identify $20$ clusters in this dataset. Table \ref{tab:a2} presents the results for dataset A2 with $35$ clusters, and again, all indices correctly indicate the presence of $35$ clusters. Finally, Table \ref{tab:a3} summarizes the results for dataset A3, which includes $50$ clusters. In this case, all indices -- except the $Dn$ index -- confirm the existence of $50$ clusters. The $Dn$ index instead suggests the presence of $45$ or $46$ clusters.

\begin{table}[h]
	\centering\footnotesize
     \renewcommand{\arraystretch}{0.75} 
	\begin{threeparttable}
		\resizebox{0.9\textwidth}{!}{
			\begin{tabular}{lrrrrrrrr}
				\toprule
				$k$ &   $T_k$   &$S_{av}$&  $DB$ & $XB$  & $Dn$  &$CH^{\natural}$&$G_{str}$&$G_{rex}$ \\
				\midrule
				17  &  3.018    & 0.557  & 0.577 & 0.087 & 0.009 &   1.411       & -0.251  &  1.583   \\
				18  &  2.691    & 0.575  & 0.537 & 0.082 & 0.010 &   1.433       & -0.153  &  1.850    \\
				19  &  2.708    & 0.585  & 0.533 & 0.079 & 0.010 &   1.460       & -0.144  &  1.879    \\
				20  &{\bf 2.466}&{\bf 0.595} &{\bf 0.527} &{\bf 0.079}&{\bf 0.023}&{\bf 1.488} &{\bf -0.090}&{\bf 2.030} \\
				21  &  2.585    & 0.583  & 0.572 & 0.271 & 0.017 &   1.436       & -0.118  &  1.972    \\
				22  &  2.654    & 0.567  & 0.616 & 0.259 & 0.017 &   1.384       & -0.163  &  1.878    \\
				23  &  2.778    & 0.551  & 0.667 & 0.293 & 0.017 &   1.337       & -0.224  &  1.754    \\
				\bottomrule
		\end{tabular}}
    \begin{tablenotes} 
	\scriptsize
    \vspace{-1mm}
	\item[$\natural$] $\times 10^{3}$
	\end{tablenotes}
	\end{threeparttable}\vspace{-5mm}
	\caption{Validity indices for different values of $k$ - dataset A1.}
	\label{tab:a1}
\end{table}

\begin{table}[h]
	\centering\footnotesize
     \renewcommand{\arraystretch}{0.75} 
	\begin{threeparttable}
		\resizebox{0.9\textwidth}{!}{
			\begin{tabular}{lrrrrrrrr}
				\toprule
				$k$ &$T_k$&$S_{av}$&  $DB$ & $XB$  & $Dn$  & $CH^{\natural}$ & $G_{str}$ & $G_{rex}$ \\
				\midrule
				32  & 2.519   &  0.580 & 0.533 & 0.081 & 0.008 & 1.683 & -0.132 & 1.866 \\
				33  & 2.535   &  0.583 & 0.536 & 0.077 & 0.008 & 1.703 & -0.145 & 1.909 \\
				34  & 2.553   &  0.591 & 0.529 & 0.071 & 0.008 & 1.728 & -0.123 & 2.004 \\
				35  &{\bf 2.438}& {\bf 0.598} & {\bf 0.524} &{\bf 0.071} & {\bf 0.010}&{\bf 1.752} &{\bf -0.089}&{\bf 2.100} \\
				36  & 2.503   &  0.591 & 0.551 & 0.249 & 0.010 & 1.718 & -0.105 & 2.065 \\
				37  & 2.539   &  0.582 & 0.577 & 0.243 & 0.010 & 1.682 & -0.130 & 2.012 \\
				38  & 2.603   &  0.572 & 0.609 & 0.280 & 0.010 & 1.647 & -0.165 & 1.943 \\
				\bottomrule
		\end{tabular}}
 \begin{tablenotes} 
		\scriptsize
        \vspace{-1mm}
		\item[$\natural$] $\times 10^{3}$
	\end{tablenotes}        
	\end{threeparttable}\vspace{-5mm}
	\caption{Validity indices for different values of $k$ - dataset A2.}
	\label{tab:a2}
\end{table}

\begin{table}[h]
	\centering\footnotesize
    \renewcommand{\arraystretch}{0.75} 
	\begin{threeparttable}
		\resizebox{0.9\textwidth}{!}{
			\begin{tabular}{lrrrrrrrr}
				\toprule
				$k$ & $T_k$   &$S_{av}$&  $DB$ & $XB$  & $Dn$  &$CH^{\natural}$&$G_{str}$&$G_{rex}$ \\
				\midrule
				45  &  2.492    & 0.578     & 0.534      & 0.088     & 0.010     &   1.966   &  -0.133   &  1.848     \\
				46  &  2.499    & 0.582     & 0.532      & 0.083     &{\bf 0.010}&   1.980   &  -0.130   &  1.861     \\
				47  &  2.473    & 0.589     & 0.527      & 0.073     & 0.008     &   2.000   &  -0.110   &  1.940     \\
				48  &  2.499    & 0.590     & 0.529      & 0.072     & 0.008     &   2.021   &  -0.120   &  1.964     \\
				49  &  2.512    & 0.596     & 0.522      & 0.069     & 0.008     &   2.045   &  -0.098   &  2.043     \\
				50  &{\bf 2.418}&{\bf 0.601}& {\bf 0.519}&{\bf 0.069}& 0.009     &{\bf 2.063}&\bf{-0.072}&{\bf 2.121} \\
				51  &  2.462    & 0.596     & 0.538      & 0.242     & 0.009     &   2.034   &  -0.083   &  2.099     \\
				52  &  2.487    & 0.590     & 0.557      & 0.238     & 0.009     &   2.004   &  -0.100   &  2.063     \\
				53  &  2.529    & 0.583     & 0.580      & 0.276     & 0.009     &   1.975   &  -0.124   &  2.015     \\
				\bottomrule
		\end{tabular}}
	\begin{tablenotes} 
		\scriptsize
        \vspace{-1mm}
		\item[$\natural$] $\times 10^3$
	\end{tablenotes}        
	\end{threeparttable}\vspace{-5mm}
	\caption{Validity indices for different values of $k$ - dataset A3.}
	\label{tab:a3}
\end{table}

\begin{figure*}[!t]
	\centering
	\subfloat[A1]{
		\includegraphics[width=0.33\textwidth]{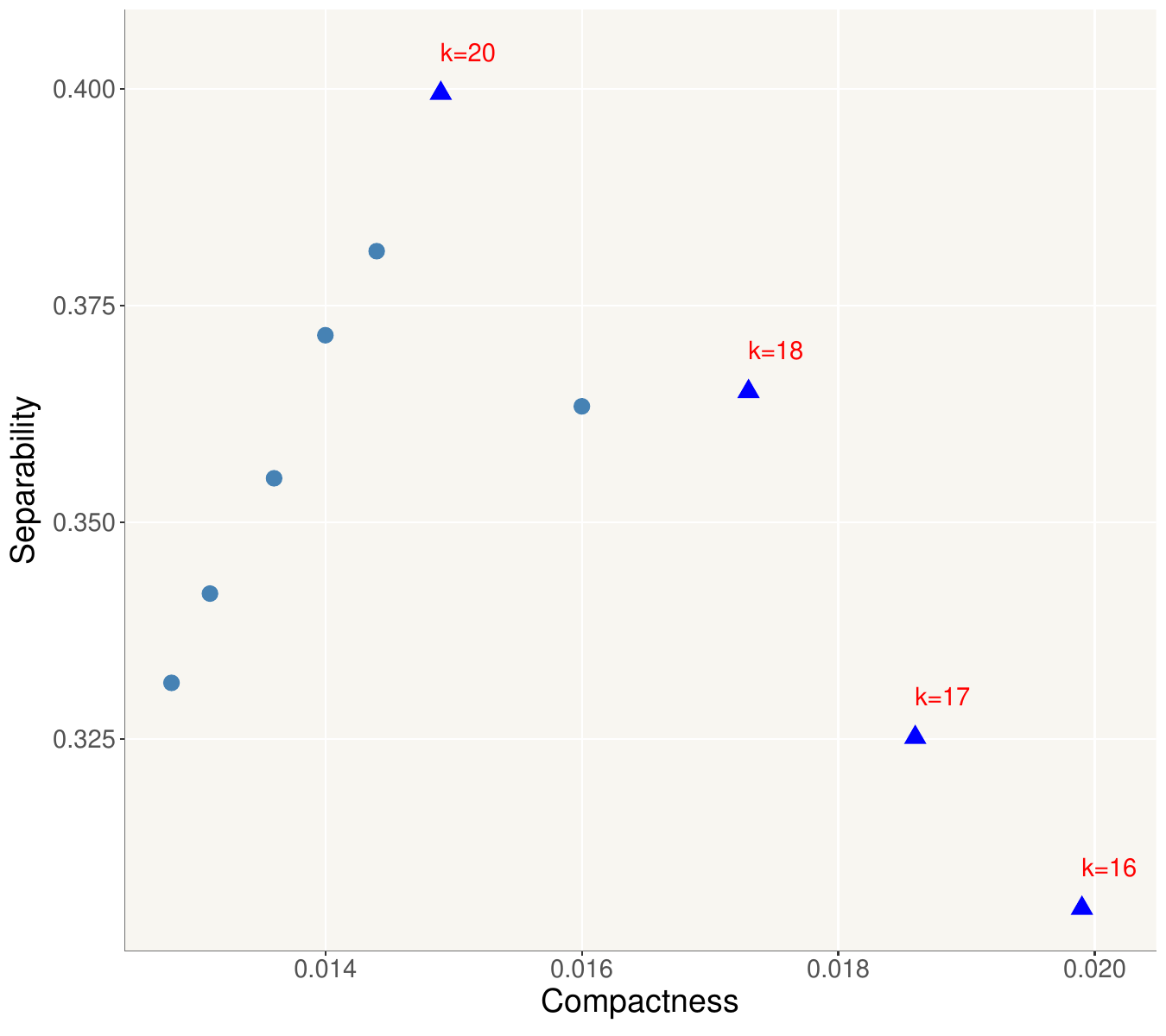}}
	\subfloat[A2]{
		\includegraphics[width=0.33\textwidth]{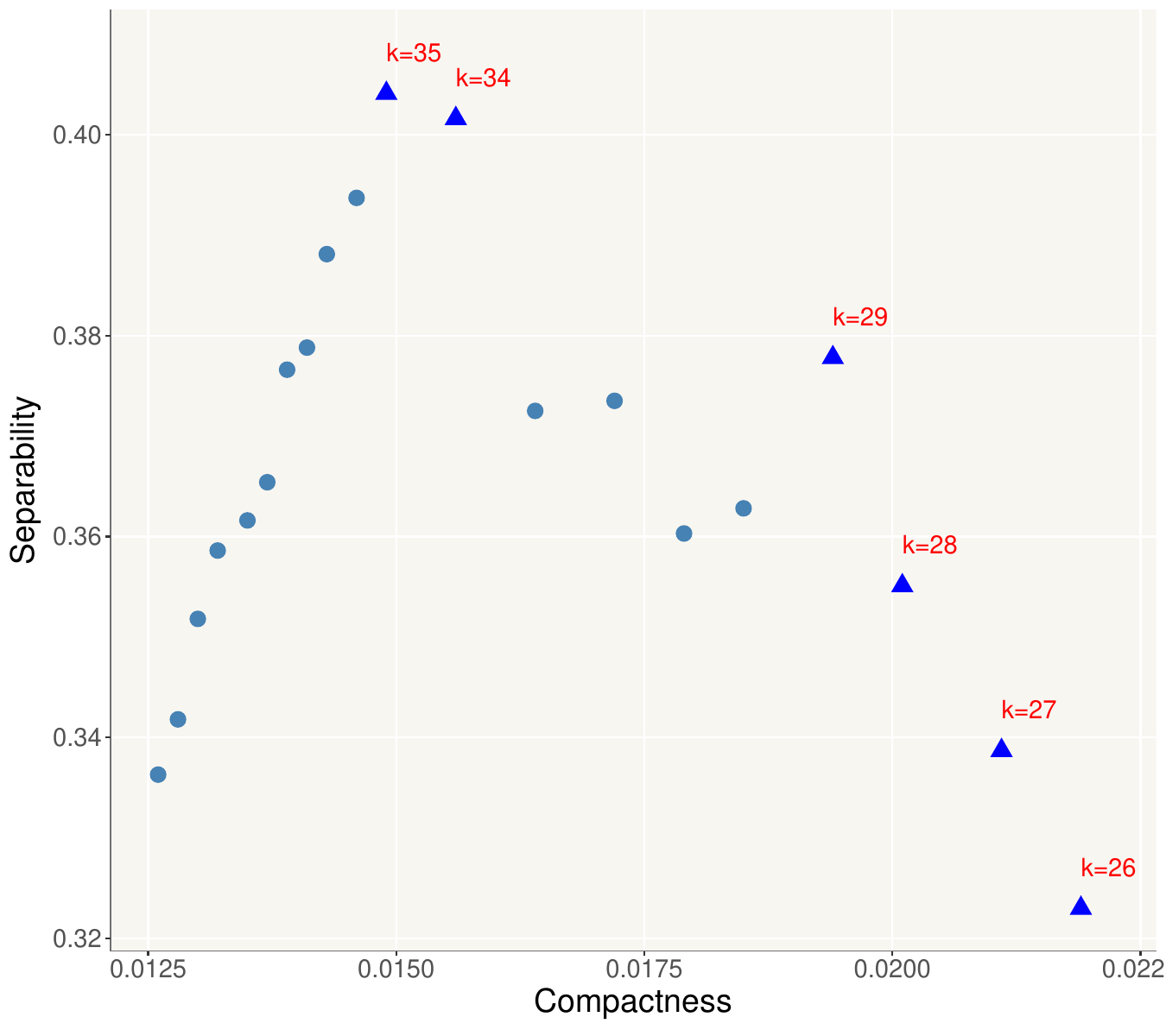}}
	\subfloat[A3]{
		\includegraphics[width=0.33\textwidth]{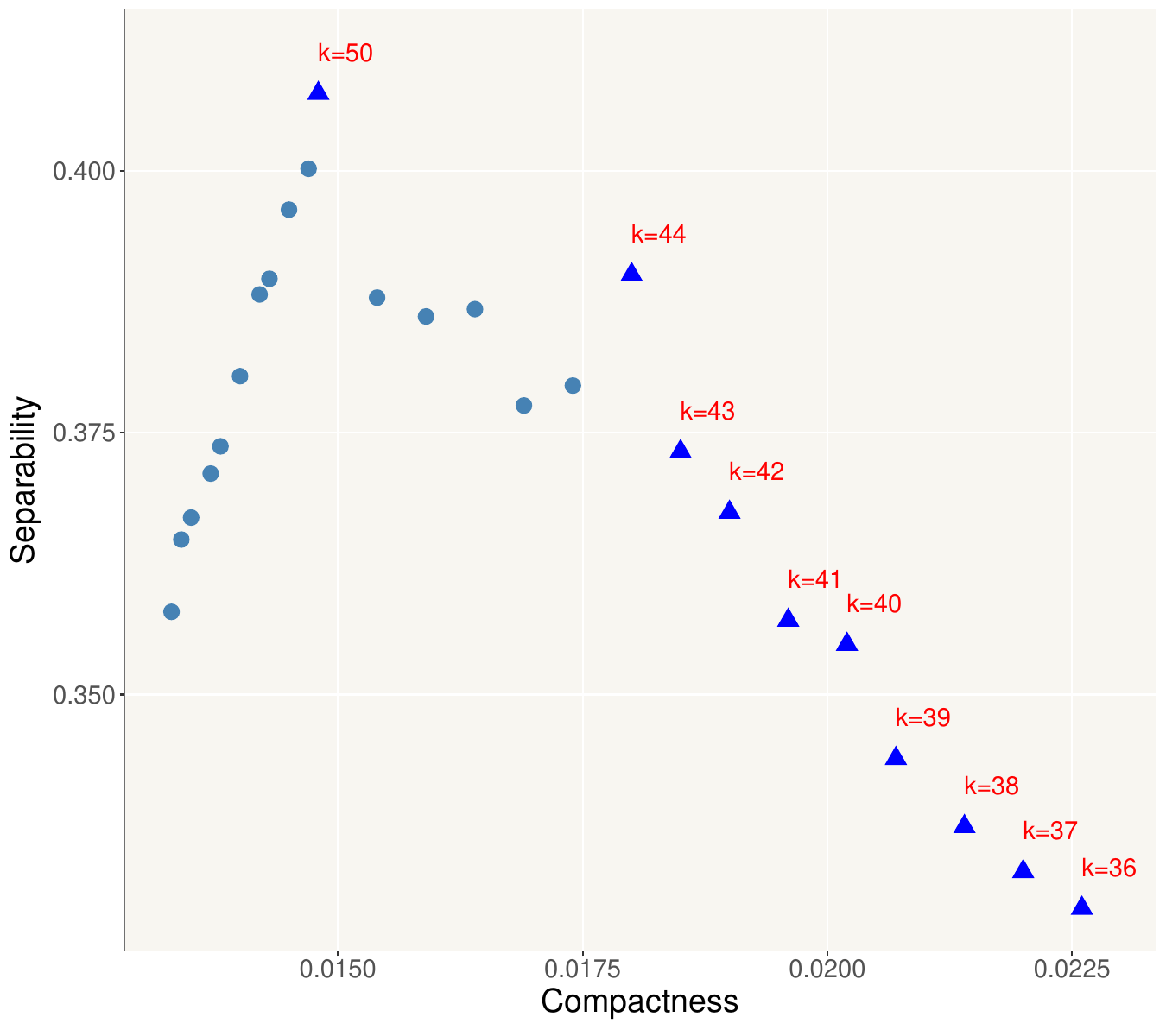}}
	\caption{Decision-space plots for A1, A2 and A3 (blue triangles indicate non-dominated solutions).}
	\label{Ds:a123}
\end{figure*}

Figure \ref{Ds:a123} shows the decision-space plots for the A1, A2, and A3 datasets. In each plot, the number of clusters associated with non‑dominated compactness and separability indices is marked with blue triangles. Across all three datasets, the true number of clusters aligns with the highest non‑dominated points.

The results of the imbalance dataset are presented in Table \ref{tab:unbalance}. This dataset contains eight clusters, and all indices confirm this, with the exception of the $CH$ index, whose value for the eight clusters is nevertheless close to its optimum. The Dim256 dataset is sparse, but its data points are distributed in 16 clusters. The corresponding results are shown in Table \ref{tab:dim256}, and all indices consistently indicate the presence of 16 clusters.

\begin{table}[h]
\centering\footnotesize
 \renewcommand{\arraystretch}{0.75} 
\begin{threeparttable}
\resizebox{0.9\textwidth}{!}{
\begin{tabular}{lrrrrrrrr}
\toprule
$k$ &$T_k$&$S_{av}$& $DB$  & $XB$  & $Dn$  &$CH^{\natural}$&$G_{str}$&$G_{rex}$ \\
\midrule
6   &  1.226  & 0.840  & 0.405 & 0.051 & 0.031 & {\bf 10.990}  &   2.049 &  5.946   \\
7   &  1.225  & 0.850  & 0.399 & 0.028 & 0.028 &   10.697      &   2.317 &  7.137   \\
8   &{\bf 1.204}&{\bf 0.858}  &{\bf 0.290} &{\bf 0.017} &{\bf 0.240}&   10.645      &{\bf 2.753} &{\bf 8.453}   \\
9   &  1.713  & 0.692  & 0.494 & 0.896 & 0.002 &    9.873      &   1.686 &  6.195   \\
10  &  2.829  & 0.528  & 0.660 & 0.827 & 0.001 &    9.315      &   0.633 &  3.934   \\
\bottomrule
     \end{tabular}}
     \begin{tablenotes} 
            \scriptsize
            \vspace{-1mm}
            \item[$\natural$] $\times 10^3$
     \end{tablenotes}     
     \end{threeparttable}\vspace{-5mm}
\caption{Validity indices for different values of $k$ - Unbalance dataset}
\label{tab:unbalance}
\end{table}

\begin{table}[h]
\centering\footnotesize
 \renewcommand{\arraystretch}{0.75} 
\begin{threeparttable}
\resizebox{0.9\textwidth}{!}{
\begin{tabular}{lrrrrrrrr}
\toprule
$k$ & $T_k$   & $S_{av}$&  $DB$ & $XB$  & $Dn$  &$CH^{\natural}$&$G_{str}$&$G_{rex}$ \\
\midrule
14  &  0.697    & 0.858     &   0.677   &  0.018    & 0.912      &    0.447  &   1.681    &   2.286      \\
15  &  0.677    & 0.922     &   0.546   &  0.010    & 0.976      &    0.722  &   4.815    &   6.031      \\
16  &{\bf 0.629}&{\bf 0.983}&{\bf 0.025}&{\bf 0.000}&{\bf 14.850}&{\bf 4.357}&{\bf 38.582}&{\bf 104.464} \\
17  &  0.631    & 0.951     &   0.068   &  0.143    & 0.532      &    4.093  &   36.421   &  98.527      \\
18  &  0.632    & 0.925     &   0.098   &  0.187    & 0.468      &    3.858  &   34.172   &  92.241      \\
\bottomrule
\end{tabular}}
\begin{tablenotes} 
       \scriptsize
        \vspace{-1mm}
       \item[$\natural$] $\times 10^3$
\end{tablenotes}
\end{threeparttable}\vspace{-5mm}
\caption{Validity indices for different values of $k$ - Dim256 dataset}
\label{tab:dim256}
\end{table}

The results for the DA1, DA2, and DA3 datasets, shown in Figure \ref{marmatrix}, are summarized in Table \ref{tab:synthetic}. Each of these datasets contains four clusters. In DA1, the clusters are well separated and all indices correctly identify four clusters. In DA2, the clusters are positioned closer to each other, but remain distinguishable. For this dataset, the $S_{av}$ and $XB$ indices suggest the presence of three clusters, while the $CH$ index indicates two clusters. All remaining indices correctly identify four clusters.

\begin{figure*}[!t]
	\centering
	\subfloat[DA1]{
		\includegraphics[width=0.33\textwidth]{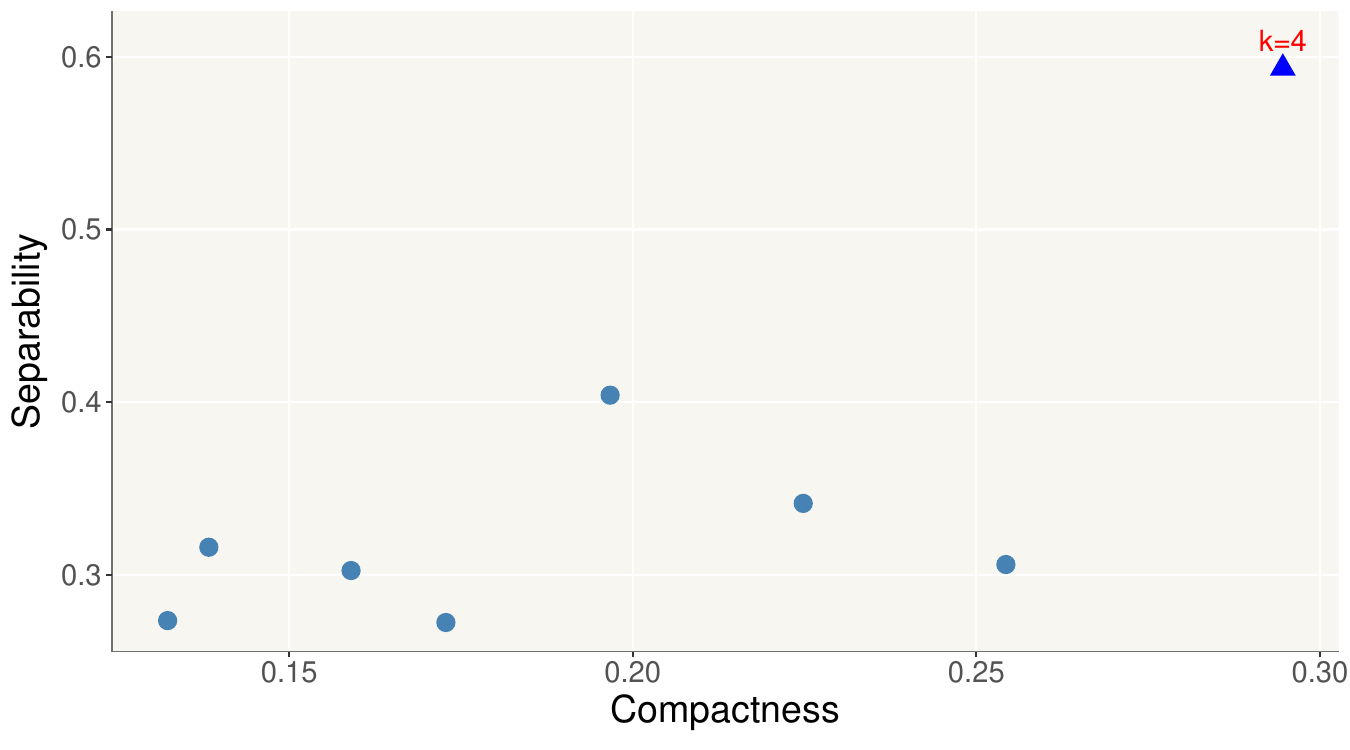}}
	\subfloat[DA2]{
		\includegraphics[width=0.33\textwidth]{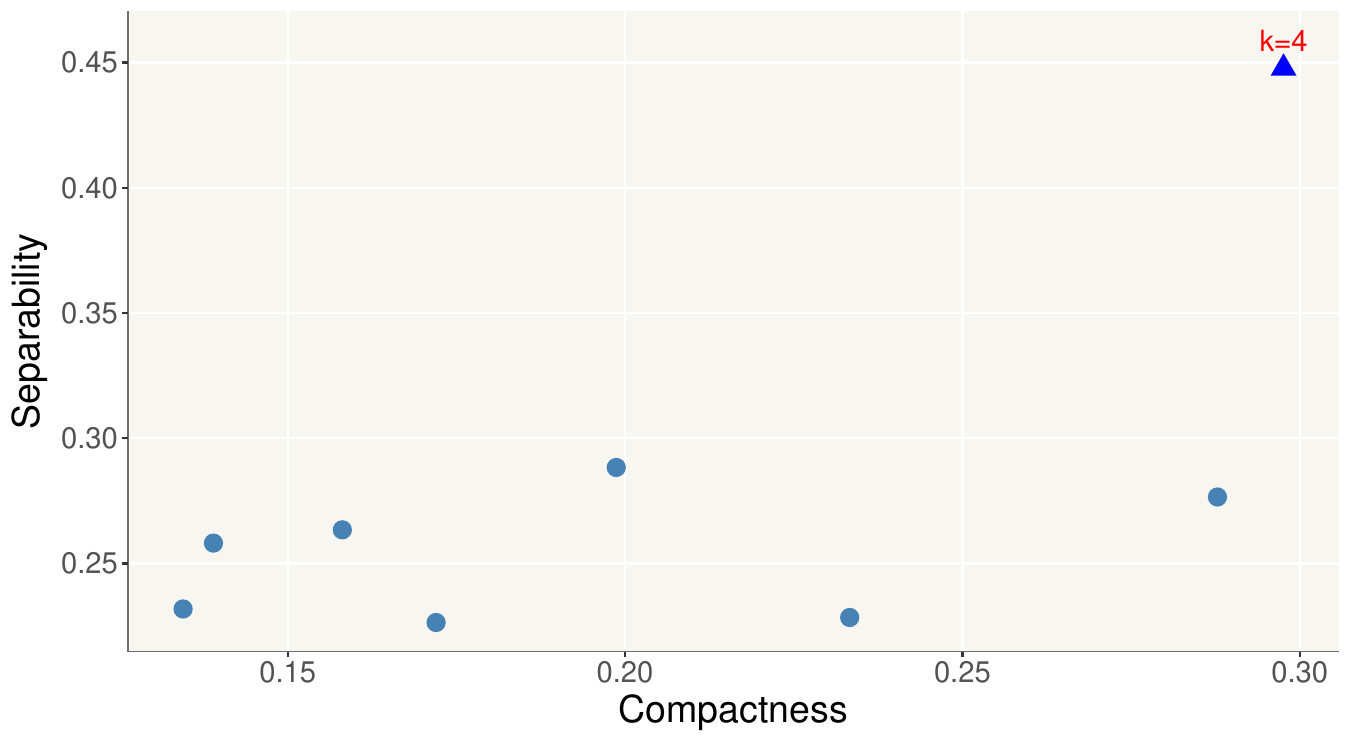}}
	\subfloat[DA3]{
		\includegraphics[width=0.33\textwidth]{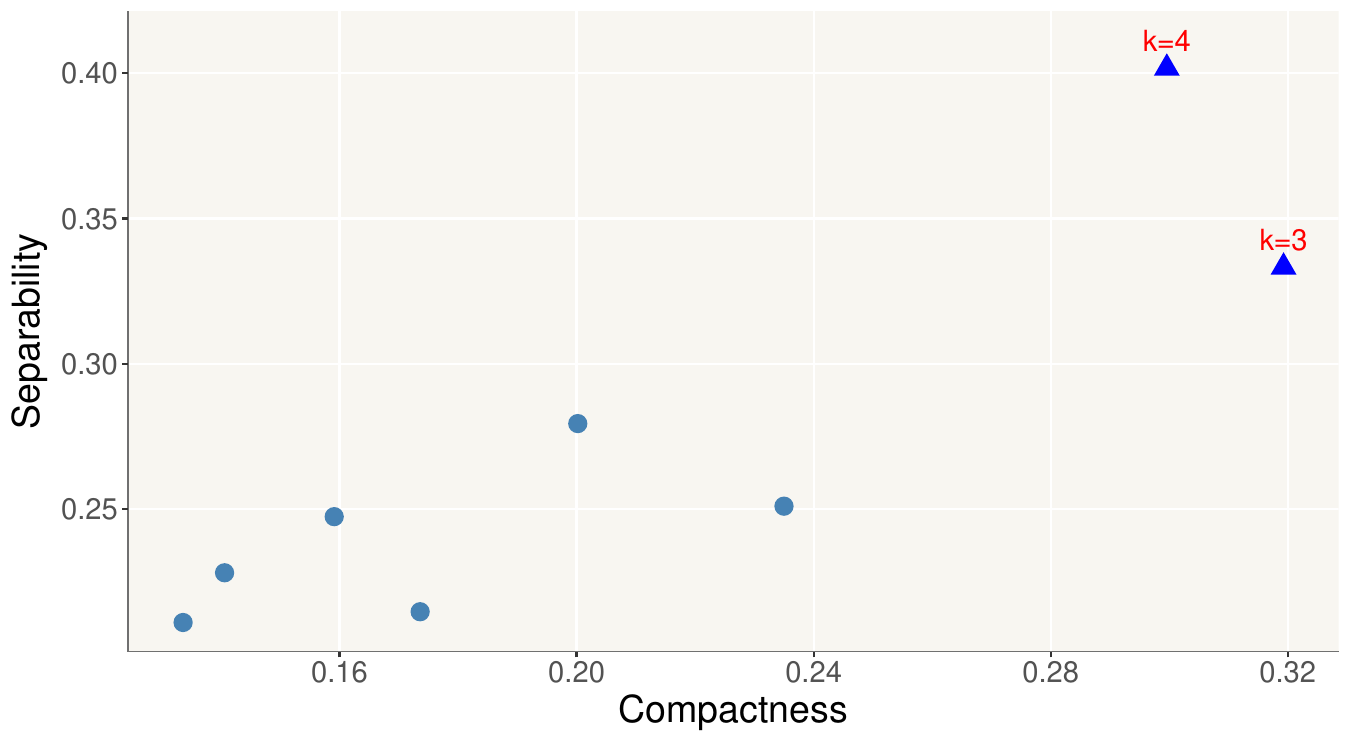}}
	\caption{Decision-space plots for synthetic datasets with 4 clusters.}
	\label{Ds1ds3}
\end{figure*}

In the DA3 dataset, the central cluster is not separated from the other three. In this case, only the $T_k$ and $Dn$ indices detect four clusters. All other indices -- except for $CH$ -- indicate three clusters by effectively partitioning the central cluster among the others. Figure \ref{Ds1ds3} displays the decision-space plots for these three datasets. In all cases, the $k$-partitions with $k=4$ have the highest separability among the non-dominated partitions.

\begin{table}[h]
\centering\footnotesize
 \renewcommand{\arraystretch}{0.75} 
\begin{threeparttable}
\resizebox{0.9\textwidth}{!}{
\begin{tabular}{lrrrrrrrr}
\toprule
$k$ &$T_k$&$S_{av}$&  $DB$ & $XB$  & $Dn$  &$CH^{\natural}$&$G_{str}$&$G_{rex}$ \\
\midrule
\multicolumn{9}{c}{\textbf{Data with well-separated clusters (DA1)}} \\
\cmidrule(lr){2-9}
2 & 1.760 & 0.645 & 0.445 & 0.083 & 0.022 & 2.178 & 0.093 & 2.699  \\
3 & 2.438 & 0.748 & 0.501 & 0.046 & 0.047 & 2.660 & 0.381 & 2.900  \\
4 &{\bf 1.189} & {\bf 0.774} &{\bf 0.322} &{\bf 0.034} &{\bf 0.344} & {\bf 2.673} &{\bf 1.097} &{\bf 4.674} \\
5 & 2.271 & 0.549 & 0.692 & 0.481 & 0.007 & 2.213 & 0.081 & 2.622  \\
6 & 1.988 & 0.563 & 0.620 & 0.344 & 0.005 & 1.931 & 0.298 & 3.027  \\
\midrule
\multicolumn{9}{c}{\textbf{Data with closer clusters but still distinct (DA2)}} \\
\cmidrule(lr){2-9}
2 & 1.693 & 0.647 & 0.534 & 0.082 & 0.011 &{\bf 2.688} &  0.071 & 1.875 \\
3 & 2.576 & {\bf 0.679} & 0.560 &{\bf 0.065} & 0.013 & 2.280 & -0.072 & 2.324 \\
4 &{\bf 1.569} & 0.671 &{\bf 0.486} & 0.072 &{\bf 0.046} & 1.905 &{\bf 0.138} &{\bf 2.638} \\
5 & 3.357 &  0.480 & 0.803 & 0.428 & 0.007 & 1.582 & -0.566 & 1.197 \\
6 & 2.780 &  0.492 & 0.713 & 0.298 & 0.005 & 1.380 & -0.408 & 1.512 \\
\midrule
\multicolumn{9}{c}{\textbf{Data with mixed clusters (DA3)}} \\
\cmidrule(lr){2-9}
2 & 5.558 & 0.586 & 0.650 & 0.111 & 0.019 &{\bf 2.236} & -0.341 & 1.247  \\
3 & 2.042 & {\bf 0.660} &{\bf 0.535} &{\bf 0.061} & 0.015 & 2.091 &{\bf -0.114} &{\bf 2.190}  \\
4 &{\bf 1.743}& 0.627 & 0.550 & 0.092 &{\bf 0.027} & 1.657 & -0.131 & 2.007  \\
5 & 3.047 &  0.468 & 0.827 & 0.418 & 0.006 & 1.387 & -0.618 & 1.029  \\
6 & 2.861 &  0.477 & 0.734 & 0.286 & 0.005 & 1.207 & -0.460 & 1.352  \\
		\bottomrule
		\end{tabular}}
    \begin{tablenotes} 
		\scriptsize
        \vspace{-1mm}
		\item[$\natural$] $\times 10^3$
	\end{tablenotes}        
	\end{threeparttable}\vspace{-5mm}
	\caption{Validity indices for different values of $k$ - Synthetic datasets with 4 clusters.}
	\label{tab:synthetic}
\end{table}

\begin{figure*}[!t] 
	\centering
	\subfloat[Liver Disorders]{
		\includegraphics[width=0.33\textwidth]{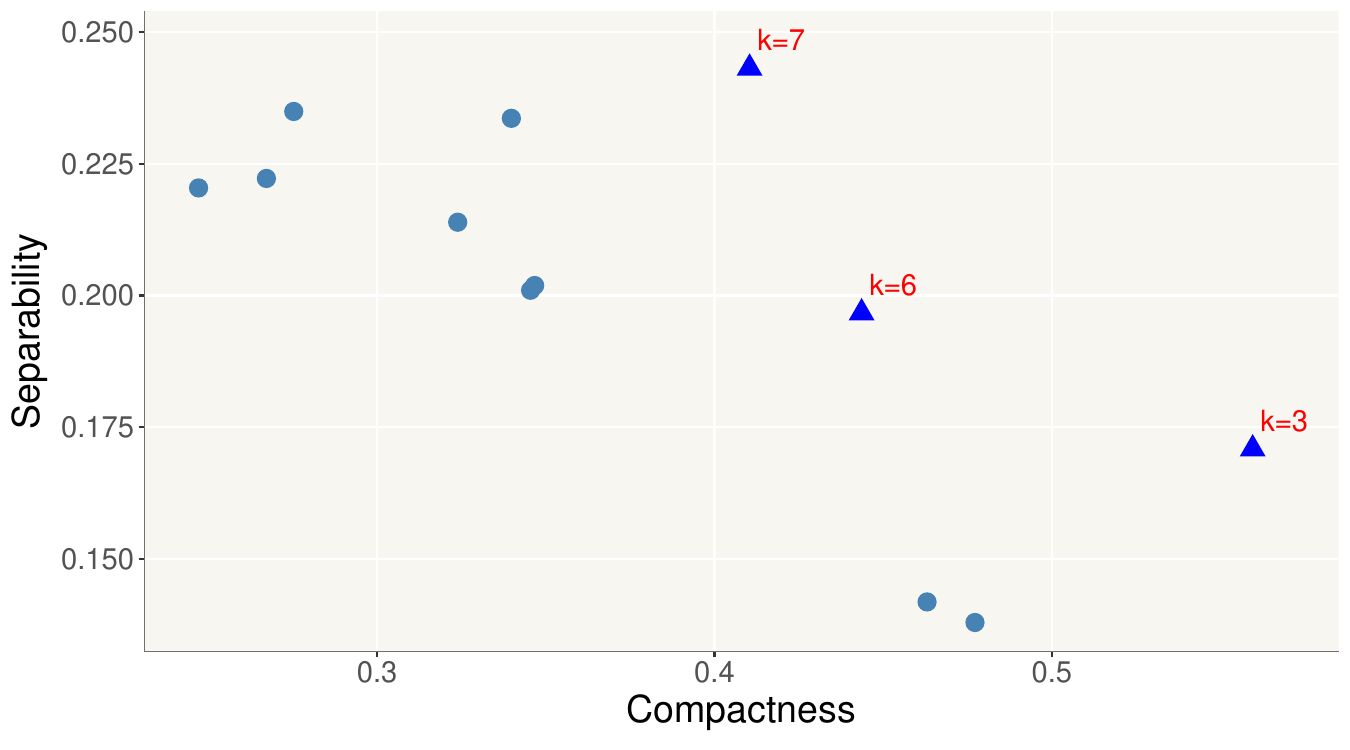}}
	\subfloat[Ionosphere]{
		\includegraphics[width=0.33\textwidth]{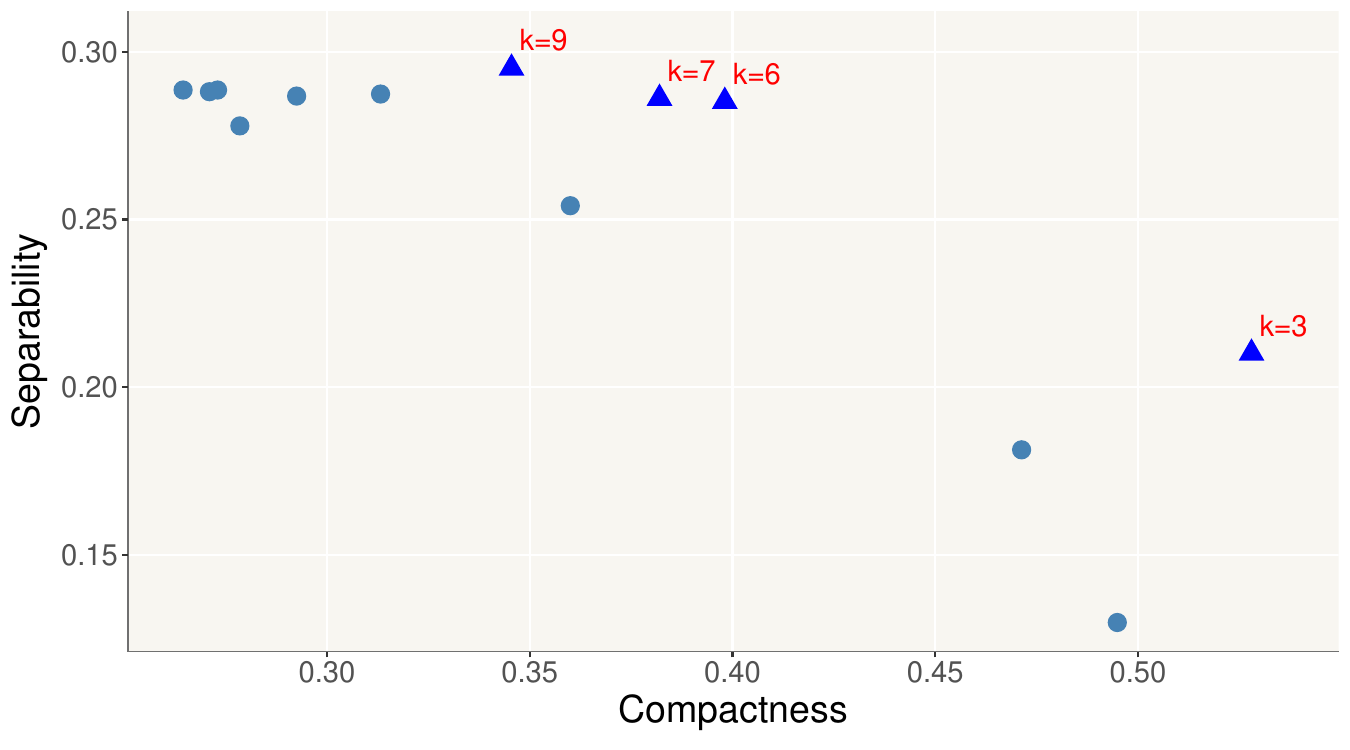}}
	\subfloat[Land Satellite]{
		\includegraphics[width=0.33\textwidth]{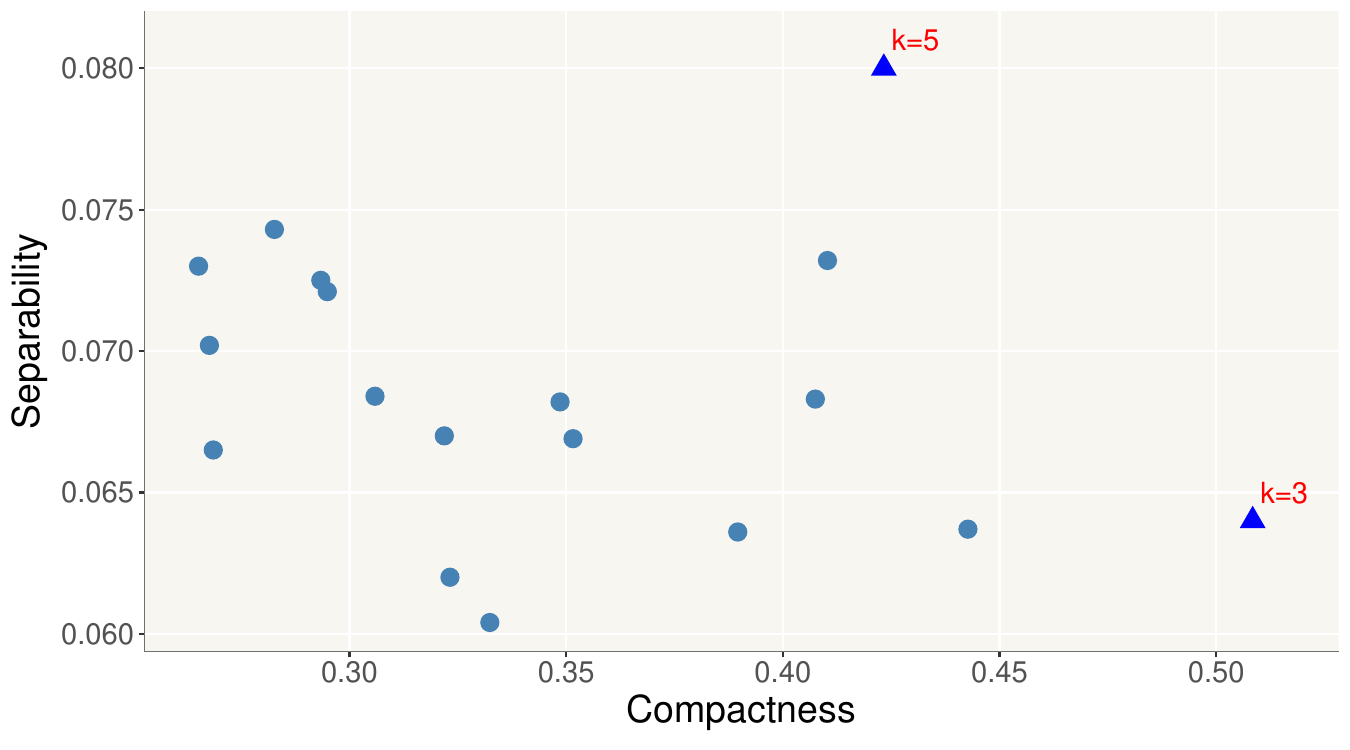}}
	\caption{Decision-space plots for Liver Disorders, Ionosphere and Land Satellite datasets.}
	\label{LDILS}
\end{figure*}

Decision-space plots for the Liver Disorders, Ionosphere, and Land Satellite datasets are shown in Figure \ref{LDILS}. These plots indicate the presence of seven clusters in the Liver Disorders dataset, nine clusters in the Ionosphere dataset, and five clusters in the Land Satellite dataset. The corresponding $k$-partitions represent the highest-ranked non-dominated solutions within their respective decision-space plots.

\begin{figure*}[!t]
	\centering
	\subfloat[TSPLIB3038]{
		\includegraphics[width=0.33\textwidth]{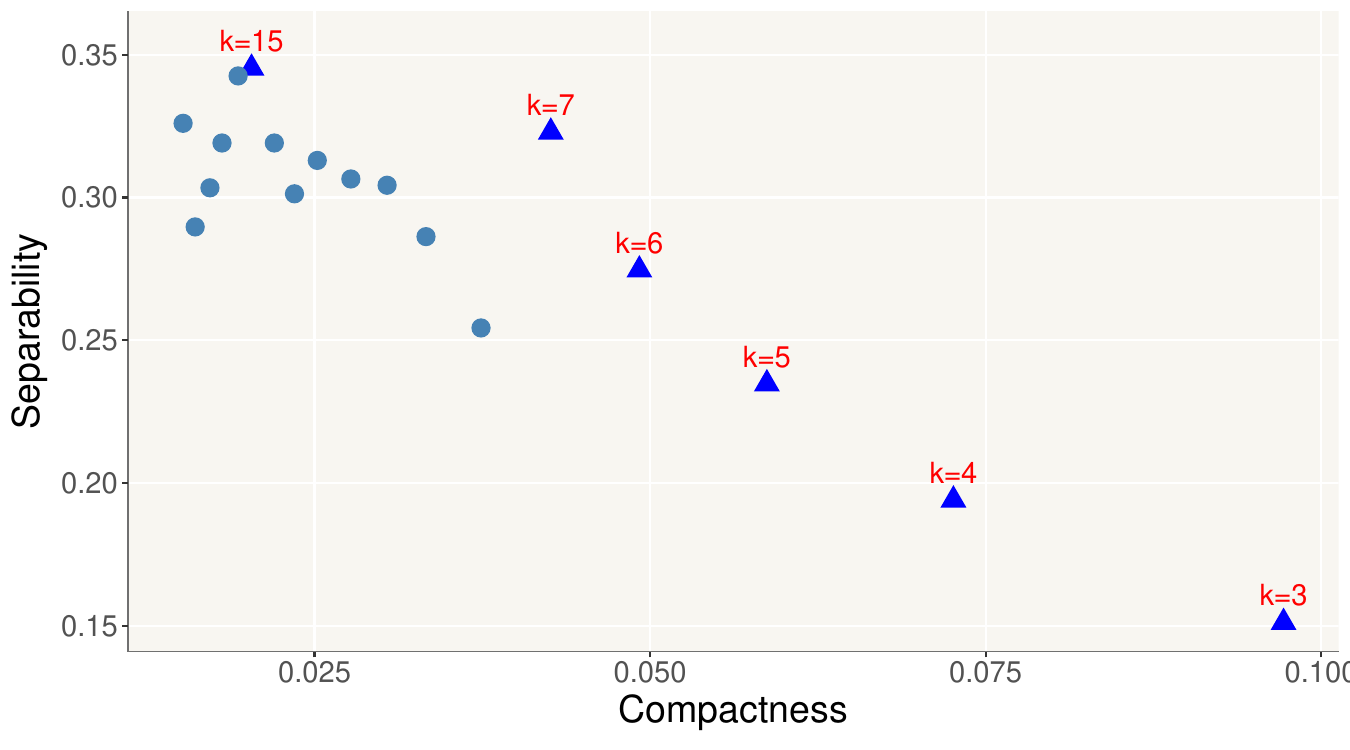}}
	\subfloat[D15112]{
		\includegraphics[width=0.33\textwidth]{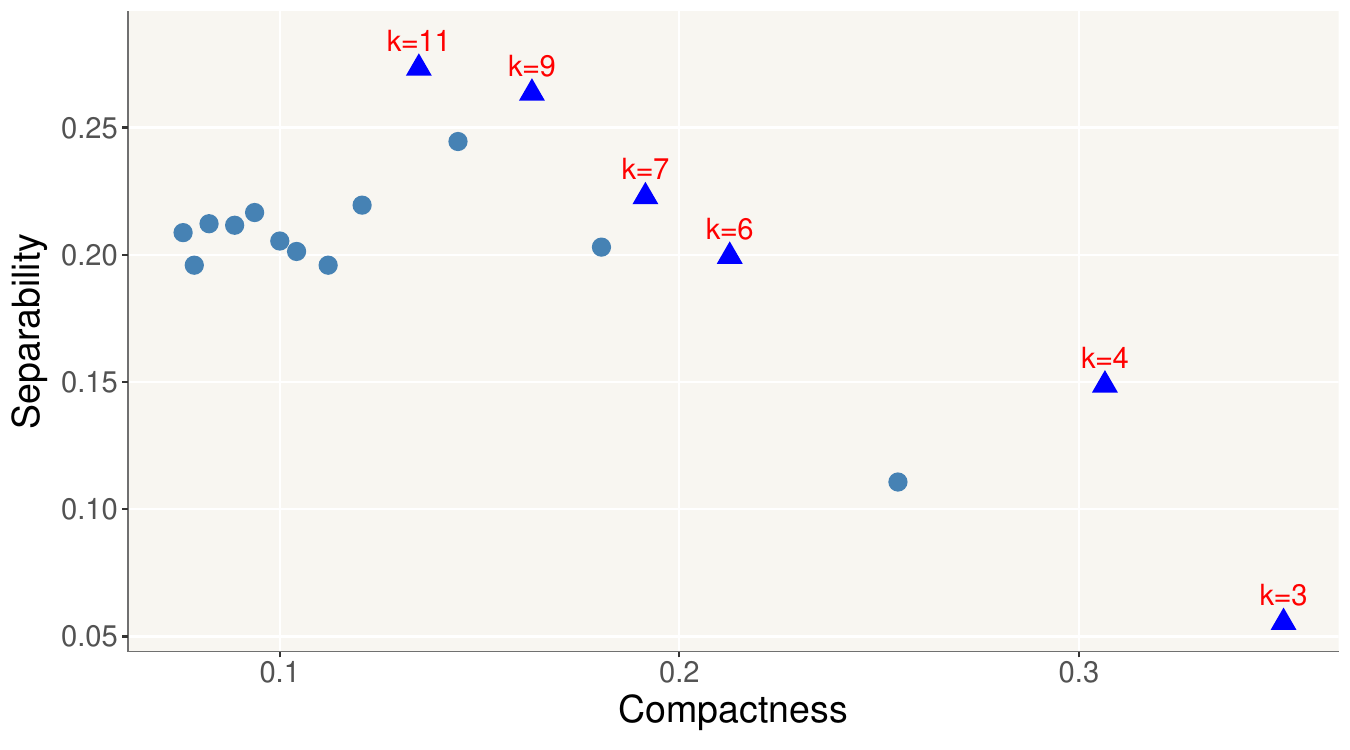}}
	\subfloat[SW24978]{
		\includegraphics[width=0.33\textwidth]{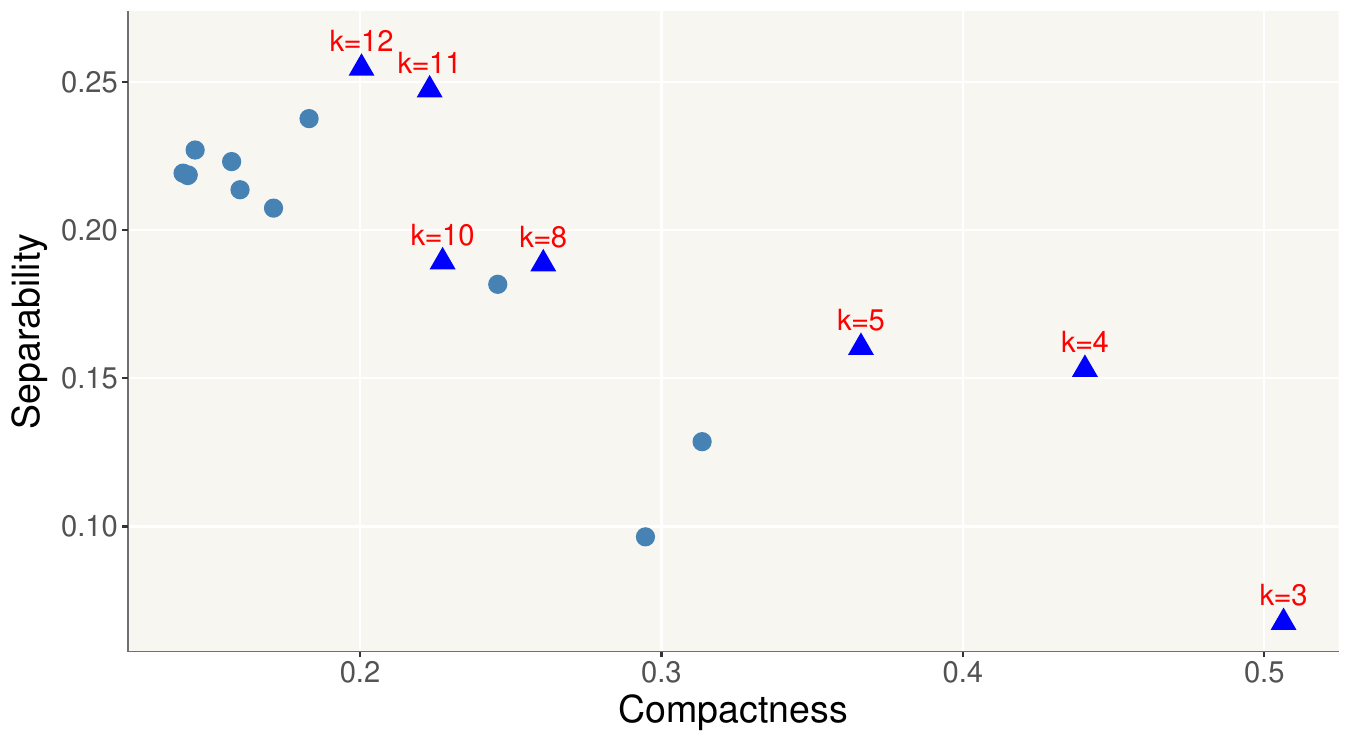}}
	\caption{Decision-space plots for TSPLIB3038, D15112 and SW24978 datasets.}
	\label{TDsw1}
\end{figure*}

Decision-space plots for three datasets - TSPLIB3038, D15112, and SW24978 - are given in Figure \ref{TDsw1}. These plots indicate that TSPLIB3038 contains 15 clusters, with seven clusters appearing as a viable alternative to the optimal cluster count.  D15112 exhibits 11 clusters, with nine clusters as an alternative, while SW24978 shows 12 clusters, with 11 clusters as an alternative. The $k$-partitions associated with these values correspond to the highest-ranking solutions among non-dominated points in the decision-space plots.

Tables \ref{tab:ShuttleControl} and\ref{tab:PersonActivity} present the results for two real-world datasets: the Shuttle Control dataset and the Localization Data for Person Activity dataset. The true number of clusters in these datasets is unknown. For Shuttle Control, there is strong agreement among the $T_k$, $S_{av}$, $XB$, $Dn$, $G_{str}$, and, to some extent, the $G_{rex}$ indices, all indicating the presence of seven clusters. For Localization Data, six indices: $T_k$, $S_{av}$, $DB$, $Dn$, $G_{str}$, and $G_{rex}$ identify 11 clusters, while the remaining indices produce notably different results. Note that this dataset contains 11 classes.

\begin{table}[h]
\centering\footnotesize
 \renewcommand{\arraystretch}{0.75} 
\begin{threeparttable}
\resizebox{0.9\textwidth}{!}{
\begin{tabular}{lrrrrrrrr}
\toprule
$k$ &  $T_k$  &$S_{av}$&  $DB$ & $XB$  & $Dn$  &$CH^{\natural}$&$G_{str}$&$G_{rex}$ \\
\midrule
6   &  1.602  & 0.980  & {\bf 0.309} & 0.001 & 0.026 &  1.712 &  1.532 & {\bf 70.610}  \\
7   &  {\bf 1.602}  & {\bf 0.981}  & 0.329 & {\bf 0.001} & {\bf 0.026} &  1.543 &  {\bf 2.061} & 70.247  \\
8   &  5.918  & 0.598  & 0.443 & 0.357 & 0.000 &  {\bf 7.536} & -2.244 &  1.395  \\
9   &  6.231  & 0.595  & 0.433 & 0.322 & 0.000 &  6.696 & -1.953 &  1.403  \\
10  &  6.204  & 0.592  & 0.429 & 0.292 & 0.000 &  6.060 & -1.490 &  1.401  \\
\bottomrule
     \end{tabular}}
    \begin{tablenotes}
        \scriptsize
        \vspace{-1mm}
    \item[$\natural$] $\times 10^3$
    \end{tablenotes}     
     \end{threeparttable}\vspace{-5mm}
\caption{Validity Indices for different values of $k$ - Shuttle Control dataset.}
\label{tab:ShuttleControl}
\end{table}

\begin{table}[h]
\centering\footnotesize
 \renewcommand{\arraystretch}{0.75} 
\begin{threeparttable}
\resizebox{0.90\textwidth}{!}{
\begin{tabular}{lrrrrrrrr}
\toprule
$k$ &$T_k$&$S_{av}$&  $DB$ & $XB$  &  $Dn$  &$CH^{\natural}$&$G_{str}$&$G_{rex}$ \\
\midrule
7   & 30.267  & 0.316  & 1.112 & 0.217 & 0.0005 & {\bf 52.483}  & -1.096  & -0.058   \\
8   & 29.147  & 0.317  & 1.034 & 0.209 & 0.0004 &   45.341      & -1.094  & -0.056   \\
9   & 25.040  & 0.323  & 1.002 & 0.164 & 0.0008 &   41.866      & -1.097  &  0.019   \\
10  & 18.918  & 0.326  & 0.940 & 0.195 & 0.0004 &   40.682      & -1.008  &  0.172   \\
11  &{\bf 13.612}& {\bf 0.339}&{\bf 0.909} & 0.170 &{\bf 0.0008}& 38.579 & {\bf -0.994}&{\bf 0.241} \\
12  & 15.696  & 0.308  & 1.034 & 0.218 & 0.0006 &   36.673      & -1.157  &  0.144   \\
13  & 14.851  & 0.306  & 1.031 & 0.162 & 0.0003 &   34.626      & -1.184  &  0.135   \\
14  & 13.990  & 0.307  & 0.977 & 0.156 & 0.0003 &   32.402      & -1.038  &  0.138   \\
15  & 14.408  & 0.302  & 0.985 &{\bf 0.145} & 0.0003 &   30.865 & -1.042  &  0.096   \\
\bottomrule
     \end{tabular}}
     \begin{tablenotes}
            \scriptsize
             \vspace{-1mm}
            \item[$\natural$] $\times 10^3$
     \end{tablenotes}     
     \end{threeparttable}\vspace{-5mm}
\caption{Validity indices for different values of $k$ - Localization data for Person Activity.}
\label{tab:PersonActivity}
\end{table}

\begin{figure*}[!h]
	\centering
	\subfloat[Shuttle Control]{
		\includegraphics[width=0.40\textwidth]{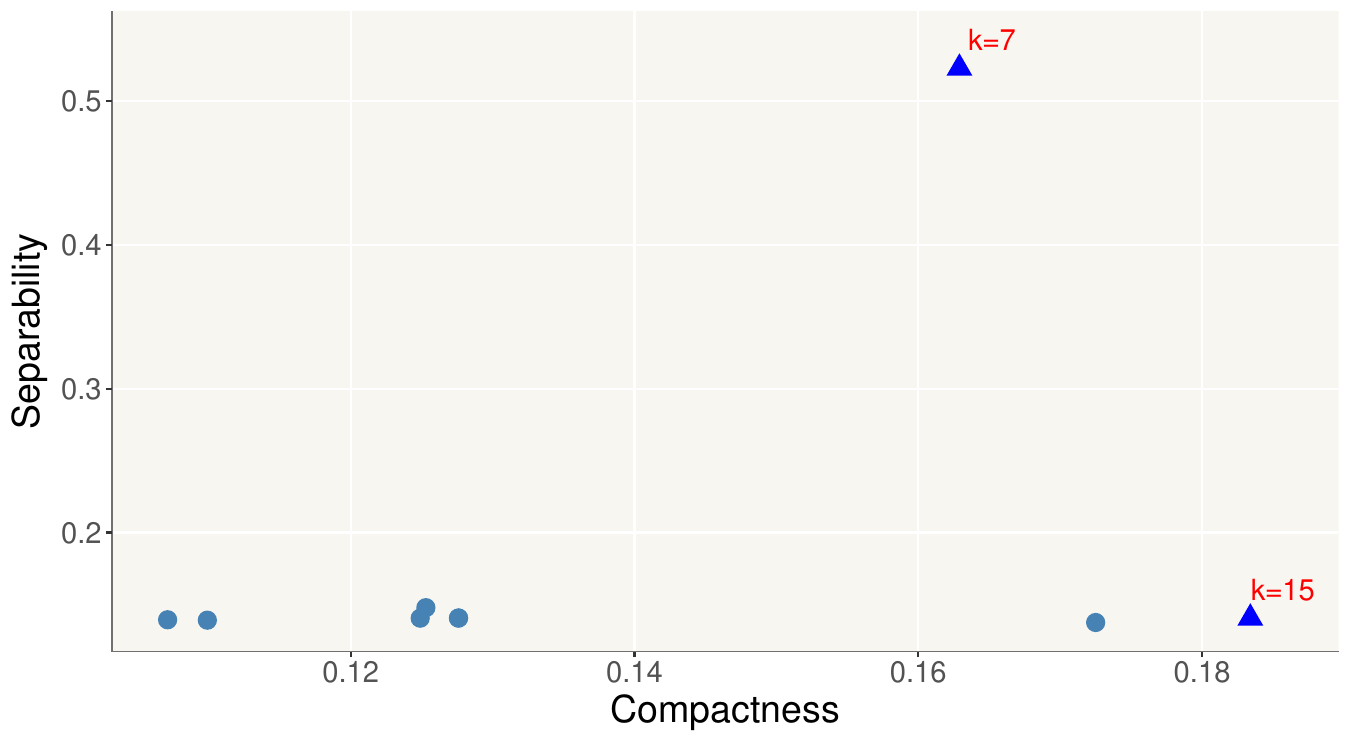}}\hspace{2mm}
	\subfloat[Person Activity]{
		\includegraphics[width=0.40\textwidth]{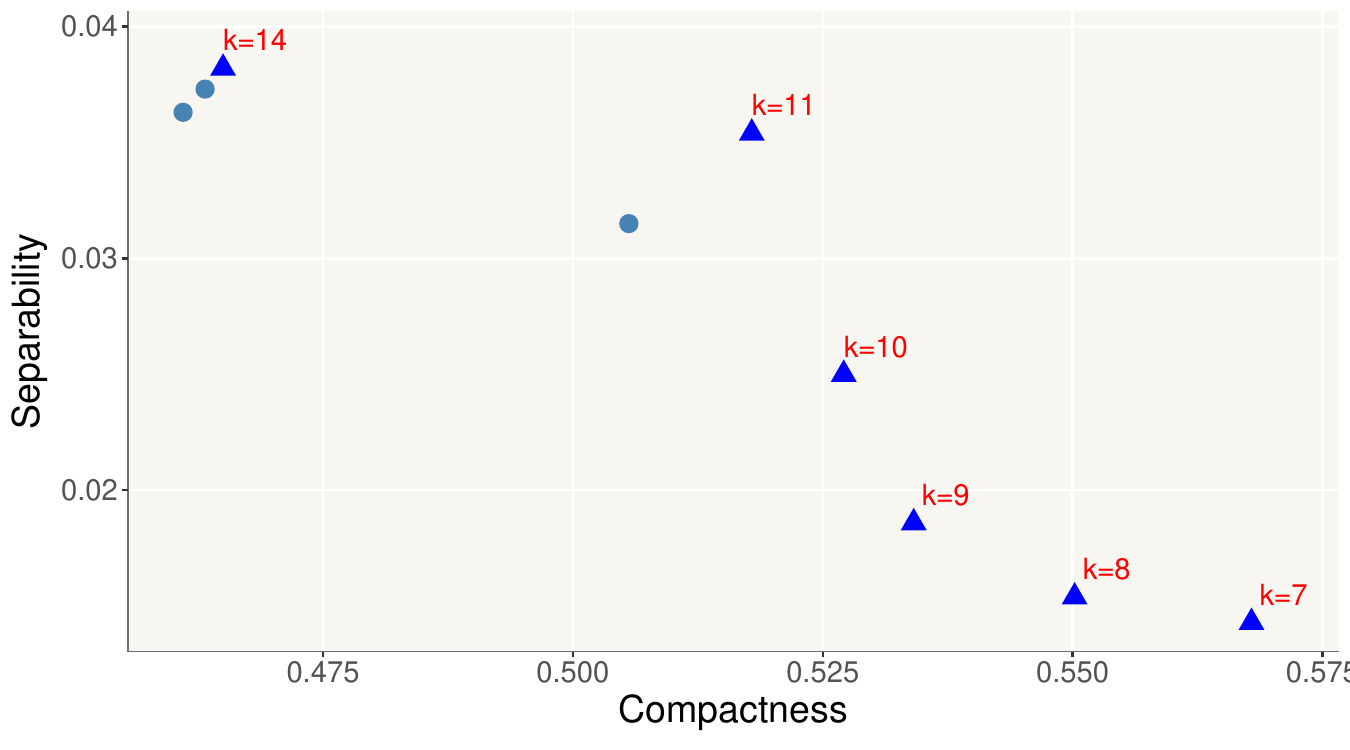}}
	\caption{Decision-space plot sets for Shuttle Control and Person Activity datasets.}
	\label{TDsw}
\end{figure*}

Decision-space plots for these datasets are provided in Figure \ref{TDsw}. These plots show that  Shuttle Control contains seven clusters, while the Localization Data exhibit 14 clusters, with 11 clusters as a reasonable alternative.

\subsection{Comparison of different cluster distributions} \label{comptwoalg}
The proposed cluster validity index, like most existing validity measures, is designed to assess the quality of clustering solutions. Since the clustering problem is inherently non-convex and admits multiple local optima, different clustering algorithms may converge to distinct solutions, resulting in varying cluster structures. Consequently, the evaluations produced by cluster validity indices can differ depending on the clustering algorithm employed. To illustrate this phenomenon, we compare the clustering solutions generated by the algorithm proposed in this paper with those obtained using the widely adopted $k$-means++ algorithm \cite{kmeansplus}. Experimental results are reported for three datasets: two synthetic datasets, A2 and Dim256, and one real-world dataset, Localization Data for Person Activity. The results are presented in Tables \ref{tab:a2-kmeans} -- \ref{tab:PersonActivity-kmeans}. For each dataset, we report the values of the clustering objective function produced by the incremental algorithm ($f_1$) and by $k$-means++ ($f_2$). Additionally, we provide the values of all considered validity indices computed for the clustering solutions obtained using $k$-means++. 

\begin{table}[h]
	\centering\footnotesize
     \renewcommand{\arraystretch}{0.75} 
	\begin{threeparttable}
		\resizebox{0.9\textwidth}{!}{
			\begin{tabular}{lrrrrrrrrrr}
			\toprule
		$k$ &$f_1^{\sharp}$ & $f_2^{\sharp}$&$T_k$&$S_{av}$& $DB$ & $XB$ & $Dn$ & $CH^{\natural}$ &$G_{str}$ & $G_{rex}$ \\
			\midrule
34&	2.241	&	3.499	&	2.720	&	0.514	&	0.690	&	0.475	&	0.011	&	1.415	&	-0.292	&	1.412 \\
35&	2.029	&	3.256	&	2.733	&	0.518	&	0.686	&	0.442	&	0.011	&	1.429	&	-0.299	&	1.438 \\
36&	1.996	&	2.907	&	2.812	&	0.530	&	0.685	&	0.411	&	0.011	&	1.465	&	-0.311	&	1.439 \\
37&	1.969	&	2.526	&	2.772	&	0.543	&	0.673	&	0.372	&	0.011	&	1.514	&	-0.283	&	1.594 \\
38&	1.942	&	2.253	&{\bf 2.608}&{\bf 0.556}&{\bf 0.652}&	0.350	&	0.011	&{\bf 1.551}&{\bf -0.224}&{\bf 1.775}\\	
39&	1.915	&	2.228	&	2.703	&	0.547	&	0.683	&{\bf 0.342}&{\bf 0.011}&	1.520	&	-0.260	&	1.685 \\
40&	1.890	&	2.208	&	2.796	&	0.536	&	0.718	&	0.401	&	0.009	&	1.489	&	-0.302	&	1.605 \\
			\bottomrule
		\end{tabular}}
 \begin{tablenotes} 
		\scriptsize
        \vspace{-1mm}
		\item[] $\sharp$ $\times 10^{10};~~$ $\natural$ $\times 10^{3}$
        \vspace{1mm}
	\end{tablenotes}        
	\end{threeparttable}\vspace{-5mm}
	\caption{Validity indices for different values of $k$ - dataset A2 - kmeans++.}
	\label{tab:a2-kmeans}
\end{table}

\begin{table}[h]
\centering\footnotesize
 \renewcommand{\arraystretch}{0.75} 
\begin{threeparttable}
\resizebox{0.9\textwidth}{!}{
\begin{tabular}{lrrrrrrrrrr}
\toprule
$k$ & $f_1^{\sharp}$ & $f_2^{\sharp}$ & $T_k$   & $S_{av}$&  $DB$ & $XB$  & $Dn$  &$CH^{\natural}$&$G_{str}$&$G_{rex}$ \\
\midrule
14&	85.197	&	94.333	&	0.692	&	0.855	&	0.708	&	0.020	&	0.845	&	0.425	&	0.987	&	2.033	\\
15&	40.451	&	39.248	&{\bf 0.677}&{\bf 0.924}&{\bf 0.521}&{\bf 0.009}&{\bf 1.006}&{\bf 0.733}&{\bf 5.288}&{\bf 6.621}\\
16&	0.241	&	39.247	&	0.685	&	0.888	&	0.709	&	31.031	&	0.023	&	0.684	&	4.964	&	6.217	\\	
17&	0.239	&	39.247	&	0.688	&	0.838	&	1.218	&   345.369 &	0.009	&	0.641	&	4.608	&	5.787	\\	
18&	0.238	&	39.246	&	0.695	&	0.790	&	1.648	&	328.454	&	0.009	&	0.602	&	4.271	&	5.392	\\
\bottomrule
\end{tabular}}
\begin{tablenotes} 
		\scriptsize
        \vspace{-1mm}
		\item[] $\sharp$ $\times 10^{6};~~$ $\natural$ $\times 10^{3}$
        \vspace{1mm}
\end{tablenotes}
\end{threeparttable}\vspace{-5mm}
\caption{Validity indices for different values of $k$ - Dim256 dataset - kmeans++}
\label{tab:dim256-kmeans}
\end{table}

\begin{table}[h]
\centering\footnotesize
\begin{threeparttable}
\resizebox{0.90\textwidth}{!}{
\begin{tabular}{lrrrrrrrrrr}
\toprule
$k$ & $f_1^{\sharp}$ & $f_2^{\sharp}$ & $T_k$   & $S_{av}$&  $DB$ & $XB$  & $Dn$  &$CH^{\natural}$&$G_{str}$&$G_{rex}$ \\
\midrule
7&	4.451	&	4.648	&{\bf 27.412}&{\bf 0.291}&	1.251	&	0.284	&{\bf 0.001}&{\bf 50.469}&	-1.234	&	-0.257	\\
8&	4.325	&	4.319	&	39.358	&	0.272	&{\bf 1.215}&{\bf 0.247}&   0.000   &45.518	&{\bf -1.189}&{\bf -0.175} \\
9&	3.967	&	4.157	&	35.374	&	0.255	&	1.299	&	0.247	&	0.000	&	40.797	&	-1.271	&	-0.287	\\
10&	3.524	&	4.011	&	29.449	&	0.240	&	1.414	&	0.312	&	0.001	&	37.275	&	-1.314	&	-0.405	\\
\bottomrule
     \end{tabular}}
\begin{tablenotes} 
		\scriptsize
        \vspace{-1mm}
		\item[] $\sharp$ $\times 10^{4};~~$ $\natural$ $\times 10^{3}$
        \vspace{1mm}
\end{tablenotes}     
     \end{threeparttable}\vspace{-5mm}
\caption{Validity indices for different values of $k$ - Localization data for Person Activity - kmeans++.}
\label{tab:PersonActivity-kmeans}
\end{table}

These tables indicate that, in most cases, the $k$-means++ algorithm converges to local solutions that are worse than those obtained by the incremental algorithm. This suggests that the clusters produced by the incremental algorithm are generally more compact than those generated by $k$-means++. As a result, cluster validity indices applied to the clustering solutions obtained by the two algorithms often yield different estimates of the number of clusters. For instance, for the A2 dataset, the indices based on the $k$-means++ solutions predict 38 or 39 clusters, whereas the dataset actually contains 35 clusters. Similarly, for the Dim256 dataset, the indices suggest the presence of 15 clusters, although the true number is 16. In the Person Activity dataset, the $k$-means++ results indicate 7 or 8 clusters, while the results reported in Table~\ref{tab:PersonActivity} support the existence of 11 clusters. These findings demonstrate that different local minima of the clustering objective function can lead to substantially different cluster configurations, which in turn may result in different estimates of the number of clusters. 

\vspace{-3mm}

\section{Discussions and conclusions} \label{conclusions}
In this paper, we introduced two indices for evaluating clustering outcomes: one that measures cluster compactness and another that assesses cluster separability. To define the compactness index, we introduced the concept of a compactness function for both the entire dataset and each cluster. This function characterizes the distribution of distances between data points and their respective cluster centers. The compactness index is then formulated with respect to a specified tolerance $\varepsilon > 0$. 

To formulate the separability index, we first introduced the concept of an adjacent set for a pair of clusters, as well as the notion of the margin between clusters based on this set. The separability index for two clusters is then defined in terms of the margin that separates them. Using the cluster centers, we next constructed a scaled version of this separability index. Finally, we extend this notion to define a separability index for the entire clustering structure.

We formulated the problem of determining the optimal number of clusters as a multiobjective optimization problem, using the compactness and separability indices as two potentially conflicting objectives -- an issue that commonly arises in real-world datasets. Within this framework, we introduced the concept of decision-space plots. A decision-space plot consists of a set of points in a two-dimensional plane, where each point represents a clustering solution with a particular number of clusters, and its coordinates correspond to the compactness and separability indices of that solution. From this plot, we identified the set of non-dominated points, and the point with the highest separability value within this set is taken to represent the optimal number of clusters. Other non-dominated points can be considered as viable alternatives to that solution. 

We assessed the proposed indices using both synthetic and real-world datasets and compared their performance with several widely used cluster validity measures. For synthetic datasets, where the true number of clusters is known, the results indicate that the proposed combined index consistently identifies the correct cluster count. The corresponding decision-space plots further show that the points representing these cluster structures have the highest separability among all non‑dominated solutions. We also evaluated the indices on eight real‑world datasets, where the findings reveal a strong alignment between the proposed indices and most of the other validity measures considered in this study.

Both the compactness and separability indices are invariant to the ordering of data points and attributes within a dataset. In addition, both indices are scaled. By selecting an appropriate value of the parameter $\varepsilon>0$, one can compute the compactness and separability for a given collection of datasets, generate decision‑space plots and compare the compactness and separability of datasets and their cluster structures. Consequently, the proposed indices function as absolute cluster validity measures. This provides an advantage over conventional indices, whose performance is often tied to specific clustering algorithms or to particular data characteristics. In addition, the use of decision-space plots offers a practical and intuitive way to examine the trade-off between two criteria, supporting not only the selection of a preferred clustering solution but also the identification of alternative, non-dominated configurations where appropriate.

Several limitations should also be acknowledged. The compactness index depends on the choice of parameters, particularly $\varepsilon$ and $\eta$. Although heuristic guidelines were provided, selecting these parameters optimally may require dataset-specific tuning. Furthermore, the approach uses cluster centers and a distance-based similarity measure, which may be less effective for datasets with highly irregular geometries or varying densities. 
Therefore, future research directions can be considered as follows. First, adaptive procedures for selecting the parameter $\varepsilon$ can be investigated to improve the robustness and automation of the proposed methodology. Large-scale implementations and parallel computing strategies could also be developed to improve computational efficiency and facilitate applications to high-dimensional and big-data environments. The applicability of the proposed indices to deep clustering and representation learning methods also warrants further investigation. Since the proposed compactness and separability indices are independent of the feature-generation mechanism, they can potentially be applied to learned latent representations. Another promising direction is the integration of the proposed compactness and separability criteria directly into clustering algorithms, allowing cluster formation and cluster validation to be performed simultaneously.

Overall, the results demonstrate that the proposed compactness and separability indices provide a meaningful and interpretable framework for assessing clustering quality and determining the number of clusters. The numerical experiments show that the approach performs competitively with established cluster validity measures while offering additional insight through its absolute interpretation and decision-space analysis. These properties make the proposed indices a promising addition to the set of tools available for cluster validation and unsupervised data analysis.


\bibliography{Abind}{}
\bibliographystyle{unsrt}

\end{document}